\documentclass[preprint,10pt,nonatbib]{elsarticle}
\usepackage[margin=1in]{geometry}
\usepackage{setspace}
\usepackage{amsmath, amssymb, graphicx, booktabs}
\usepackage[numbers,sort&compress]{natbib}
\usepackage{float}
\usepackage{subcaption}
\usepackage{threeparttable}
\usepackage{threeparttablex}
\usepackage{tikz}
\usetikzlibrary{shapes.geometric, arrows.meta, positioning, shadows,calc}
\usepackage{xcolor}
\usepackage{comment}
\usepackage{siunitx}
\usepackage{longtable}
\graphicspath{{figures_/}{figures_/Figs/}}
\usepackage{multirow}
\usepackage{caption}
\usepackage{placeins}
\usepackage{array}
\usepackage{colortbl}
\usepackage{enumitem}  % for tighter lists
\usepackage{pdfpages}

\definecolor{darkgreen}{RGB}{0,128,0}
\definecolor{medgreen}{RGB}{144,238,144}
\definecolor{medred}{RGB}{255,160,160}
\definecolor{darkred}{RGB}{200,0,0}
\definecolor{lightyellow}{RGB}{255,255,204}
\definecolor{headerblue}{RGB}{173,216,230}

\FloatBarrier

\makeatletter
\renewcommand\section{\@startsection{section}{1}{\z@}%
  {-8pt \@plus -2pt \@minus -2pt}%
  {4pt \@plus 1pt}%
  {\normalfont\large\bfseries}}
\renewcommand\subsection{\@startsection{subsection}{2}{\z@}%
  {-6pt \@plus -1pt \@minus -1pt}%
  {2pt \@plus 0.5pt}%
  {\normalfont\normalsize\bfseries}}
\renewcommand\subsubsection{\@startsection{subsubsection}{3}{\z@}%
  {-4pt \@plus -1pt \@minus -1pt}%
  {2pt \@plus 0.5pt}%
  {\normalfont\normalsize\itshape}}
\makeatother

\makeatletter
\renewcommand\paragraph{\@startsection{paragraph}{4}{\z@}%
  {4pt \@plus 1pt \@minus 1pt}%
  {-1em}%
  {\normalfont\normalsize\bfseries}}
\makeatother

\setlist[enumerate]{topsep=2pt, itemsep=1pt, parsep=0pt, leftmargin=*}
\setlist[itemize]{topsep=2pt, itemsep=1pt, parsep=0pt, leftmargin=*}

\renewcommand{\arraystretch}{1.1}

\makeatletter
\def\ps@pprintTitle{\let\@oddhead\@empty\let\@evenhead\@empty\def\@oddfoot{\hfil\thepage\hfil}\let\@evenfoot\@oddfoot}
\makeatother
\begin{document}
\begin{frontmatter}

\title{Evaluating AlphaEarth Embeddings for Crop Mapping: Separability, Crop Rotations and Cross-Continental Transferability}

\author[iitk]{Priyanka Goyal\corref{cor1}}
\ead{priyankag.1@yahoo.com}

\author[iitd]{Maitreyee Sar}
\ead{maitreyeesar@gmail.com}

\author[iitk]{Adway Mitra}
\ead{adway.cse@gmail.com}

\author[tcs]{Manjira Sinha}
\ead{manjiras@gmail.com}

\cortext[cor1]{Corresponding author}

\address[iitk]{Department of Artificial Intelligence,
Indian Institute of Technology Kharagpur,
Kharagpur 721302, West Bengal, India}
\address[iitd]{Centre for Atmospheric Sciences, Indian Institute of Technology Delhi, New Delhi 110016, India}
\address[tcs]{Tata Consultancy Services Research, Kolkata, India}

\begin{abstract}
Earth observation (EO)-based crop mapping supports food security, a central aim of United Nations Sustainable Development Goal 2 (Zero Hunger), but it is unclear whether the learned vectors (embeddings) that geospatial foundation models provide for every location improve it across years and regions. We compared AlphaEarth embeddings (Google DeepMind) with compact engineered features (spectral, normalized difference vegetation index, NDVI, and harmonic; 4--6 features each) in five US counties (2017--2023) and on smallholder fields in India (rabi 2021). Each year's features were paired with the crop grown that year, and four classifiers were used. Embeddings separated look-alike crops by larger margins than the engineered features, even at matched dimensionality, and, with the combined set, were the most accurate on counties held out of training (mean Macro F1 0.82 vs. 0.37--0.45). They also achieved the highest separability of rotation classes, though the lead was small once crop identity was removed, and their class signatures remained stable in later years for the same fields. Transfer between continents showed a different pattern. In a zero-shot test with three shared crops, embeddings retained 16--28\% of their within-region accuracy, against 60--93\% for harmonic and NDVI features, and 50 labeled target fields per class used alone (Macro F1 0.64 and 0.97) did better than adding them to the source data (0.50 and 0.92). In these experiments embeddings gave the highest accuracy within a region, engineered features transferred better without local labels, and the evaluation design offers a structured way to choose features for crop mapping.
\end{abstract}

\end{frontmatter}

% \begin{figure}[!ht]
% \centering
% \includegraphics[width=0.6\textwidth]{figures_/Graphical_abstract.pdf}
% \caption{Graphical Abstract: AlphaEarth embeddings are evaluated against traditional spectral and phenological features across geometric separability, temporal dynamics, spatial robustness, and cross-regional transferability for crop mapping.}
% \end{figure}

\FloatBarrier
\section{Introduction}
\label{sec:intro}

Producing enough food while protecting land, water and soil is the central task of United Nations Sustainable Development Goal 2 (Zero Hunger), which promotes improved nutrition and sustainable agriculture \cite{un2015agenda}. Achieving this goal requires timely information on what is grown where and in what sequence, because crop maps underpin yield estimates, rotation and land-use planning, input management and agricultural statistics \cite{omia2023remote}. Earth observation (EO) can supply this information over large areas, but the quality of a crop map depends on the features that describe each field.

In the United States the USDA Cropland Data Layer (CDL) gives a crop label for every year \cite{boryan2011monitoring}. The CDL is a classification product. It provides labels but no information on crop growth, and it describes multi-year rotations only when the annual layers are combined \cite{upcott2023new}. Monitoring crop growth, and mapping crops in regions without such labels, depends on features computed from imagery. These have mostly been hand-engineered, most often vegetation indices such as the Normalized Difference Vegetation Index (NDVI), which summarize how green a field is and when it is greenest \cite{zeng2020review, li2021high}. Vegetation indices saturate in dense canopies, are sensitive to atmospheric conditions \cite{matsushita2007sensitivity, huete2002overview} and say little about management practices. Averaging pixels over fields, as in Geographic Object-Based Image Analysis (GEOBIA), reduces noise \cite{blaschke2010object, blaschke2014geographic}, but the aggregated features are still hand-engineered summaries. Time series help because a crop's path through the season separates look-alike crops better than a single image \cite{do2019detailed}, and seasonal curves (harmonic features) are widely used \cite{roy2020robust, zeng2020review}. Harmonic features, however, assume a regular yearly cycle and do not describe multi-year behavior such as rotations \cite{upcott2023new, zhou2022crop}.

Designing features by hand for every crop, region and sensor is slow, and labeled fields are scarce and costly to collect \cite{kerner2020rapid}. A newer option is to let a model learn the features. Geospatial foundation models are trained on very large archives of unlabeled EO imagery with self-supervised objectives, such as reconstructing masked parts of an image or time series. An encoder learns to compress each location into a vector that retains the information needed to reconstruct its observations. This vector is called an embedding \cite{bommasani2021opportunities, xiao2025foundation}, and models such as SatMAE and Prithvi follow this approach \cite{cong2022satmae, szwarcman2025prithvi}. AlphaEarth Foundations is an ``embedding field'' model. Its space--time--precision encoder summarizes a year of observations from many sources (optical, radar, lidar, climate, topographic and land-cover records, and geotagged text) into one embedding for every location, trained with reconstruction, consistency and text-contrastive objectives \cite{brown2025alphaearth}. The released product gives a 64-dimensional embedding, stored as 8-bit values, for every 10~m pixel and every year from 2017, so a user receives the features without handling the source imagery.

Since its release, AlphaEarth embeddings have been tested in a range of applications. In agriculture, they have been used for crop type and cropland mapping \cite{lisaius2026embedding, cai2026annual, yang2026evaluating, murakami2025within, wang2026ai}, for yield estimation \cite{fang2026application}, and in benchmarks against other inputs \cite{ma2025harvesting, neophytides2026benchmarking, shao2026benchmarking}. Outside crops, they have been applied to land cover \cite{khan2025evaluating}, tree cover and tree species \cite{zhou2026improving, wang2026evaluating}, vegetation height \cite{karim2026alphaearth} and land surface temperature \cite{liu2026endtoend}. Other studies examine the embeddings themselves \cite{benavides2026what, borocz2025first}. Closest to this study, one benchmark applies the model to agricultural downstream tasks \cite{ma2025harvesting} and one tests within- and cross-regional crop mapping for a single cool-climate upland system \cite{murakami2025within}. This paper differs in three ways. It holds the number of dimensions constant, tests whether embeddings describe crop rotations beyond crop identity, and includes a target-only baseline in a transfer experiment between two continents.

Geospatial foundation models perform well on benchmarks \cite{han2024bridging, ma2025harvesting}, but many evaluations report accuracy for a single task in a single region. That leaves open whether embeddings capture crop-relevant patterns, whether they follow crop growth patterns over years, and whether they still work in a region unlike the one where they were trained \cite{nozawa2022evaluation, tuia2016domain}. Temporal deep-learning models suggest that learned representations may capture crop trajectories better than engineered features \cite{tarasiou2023vits, tseng2023lightweight}, and classifiers trained in one region or year often degrade elsewhere \cite{hao2020transfer, tuia2016domain}, so year-to-year behavior and transfer to new regions both need direct tests.

Judging an embedding fairly takes more than one accuracy score, and four design choices can change conclusions. First, a crop label belongs to one year, so pairing it with features from other years misdescribes the field whenever the crop rotates. Second, distance measures of class separation grow with the number of dimensions regardless of whether those features carry information \cite{richards2006remote, benediktsson2015spectral}. Third, rotation measures computed on year-averaged features largely restate which crops a rotation contains. Fourth, random cross-validation rewards spatial proximity, because neighboring samples look alike \cite{foody2002status, roberts2017cross, schratz2019hyperparameter, ferraciolli2019neglecting}.

This study tests AlphaEarth embeddings against three engineered feature sets in two contrasting farming systems, mechanized large-field agriculture in the Conterminous United States (CONUS) and fragmented smallholder farming in India (mainly Uttar Pradesh). Its aim is to find where embeddings beat engineered features and where they do not. It asks three questions:
\begin{enumerate}
\item \textbf{RQ1}: Do embeddings separate crops better than engineered features, and does this hold when the number of dimensions is matched?
\item \textbf{RQ2}: Do embeddings capture how crops change from year to year and across rotations, beyond crop identity?
\item \textbf{RQ3}: Can embedding-based models maintain accuracy in a region different from the one where they were trained, and what restores it?
\end{enumerate}

\noindent The study makes three contributions, one for each question, which the paper follows in this order:
\begin{enumerate}
\item \textbf{(C1) Separability under dimension control.} Class separation is measured with five distances and scores and with accuracy on counties held out of training. Both are repeated at matched dimensionality, because distances grow with the number of dimensions.
\item \textbf{(C2) Rotational dynamics.} Rotation separability, within-class stability and accuracy on later years are measured with features paired with the crop of each year. The rotation measures are recomputed after crop identity is removed, because they otherwise restate which crops a rotation contains.
\item \textbf{(C3) Cross-continental transferability.} Models are transferred between the two continents in both directions, and a target-only baseline separates what the source region contributes from what local labels contribute.
\end{enumerate}
\noindent Overall, embeddings were far more accurate than compact engineered features within the United States, and when models moved to India a few tens of local labels mattered more than the source model. Further analyses, including the stored-label design, other classifiers, spatial block validation and robustness checks, are given in the Supplementary Material.

\FloatBarrier
\section{Materials and Methods}
\label{sec:methods}

Features are extracted in Google Earth Engine (GEE) and analyzed in Python (Fig.~\ref{fig:workflow}).

\begin{figure}[!ht]
\centering
\resizebox{\textwidth}{!}{% TikZ source of Fig. 1, used by the manuscript (\input) and by workflow.tex (standalone PDF for upload)
{\sffamily\hyphenpenalty=10000 \exhyphenpenalty=10000
\definecolor{tabc}{HTML}{F8D7B0}
\definecolor{hlc}{HTML}{E4EEFB}
\definecolor{greyc}{HTML}{F4F4F4}
\definecolor{embc}{HTML}{CC79A7}
\definecolor{spec}{HTML}{0072B2}
\definecolor{ndvc}{HTML}{009E73}
\definecolor{harc}{HTML}{E69F00}
\definecolor{comc}{HTML}{D55E00}
\tikzset{
  panel/.style={draw=black!75, rounded corners=4pt, line width=0.7pt, fill=white, anchor=north west},
  tab/.style={draw=black!75, rounded corners=3pt, fill=tabc, font=\bfseries\footnotesize, inner sep=3pt, line width=0.7pt},
  sec/.style={font=\scriptsize\itshape, text=black!55, anchor=south east},
  src/.style={draw=black!45, rounded corners=3pt, fill=hlc, align=center, font=\footnotesize, text width=3.0cm, minimum height=1.25cm, inner sep=2pt, line width=0.5pt},
  chip/.style={rounded corners=3pt, minimum width=2.9cm, minimum height=0.8cm, align=center, font=\bfseries\footnotesize, text=white, inner sep=2pt},
  ctrl/.style={align=left, font=\footnotesize, text width=3.0cm, inner sep=1pt},
  flow/.style={-{Stealth[length=2.7mm,width=2.3mm]}, line width=1.3pt, black!85, shorten <=0.5pt, shorten >=0.5pt}
}
\begin{tikzpicture}[x=1cm,y=1cm]
% column x origins (3.4 cm panels, 0.8 cm gaps): 0, 4.2, 8.4, 12.6
% ---- 1 Data
\node[panel, minimum width=3.4cm, minimum height=7.4cm] (p1) at (0,0) {};
\node[tab] at (p1.north) {1\ Data};
\foreach \k/\t in {0/{AlphaEarth embeddings\\(64-d, annual)}, 1/{Sentinel-2 Level-2A\\(spectral, NDVI, harmonic)}, 2/{USDA CDL 2017--2023\\(US crop sequence)}, 3/{AgriFieldNet\\(India, rabi 2021)}} {
  \node[src] at (1.7,{-1.4-1.62*\k}) {\t};
}
\node[sec] at ($(p1.south east)+(-0.12,0.08)$) {Sec.\ 2.1};
% ---- 2 Field-year rows
\node[panel, minimum width=3.4cm, minimum height=7.4cm] (p2) at (4.2,0) {};
\node[tab] at (p2.north) {2\ Feature Extraction};
\foreach \k/\n/\c in {0/Embedding/embc, 1/Spectral/spec, 2/NDVI/ndvc, 3/Harmonic/harc, 4/Combined/comc} {
  \node[chip, fill=\c] at (5.9,{-1.2-0.95*\k}) {\n};
}
\node[align=center, font=\footnotesize, text width=3.0cm, draw=black!45, rounded corners=3pt, fill=hlc, minimum height=1.1cm, inner sep=2pt] at (5.9,-6.35) {paired with the crop grown in the same year};
\node[sec] at ($(p2.south east)+(-0.12,0.08)$) {Sec.\ 2.2};
% ---- 3 Evaluation
\node[panel, minimum width=3.4cm, minimum height=7.4cm] (p3) at (8.4,0) {};
\node[tab] at (p3.north) {3\ Evaluation};
\foreach \k/\t/\l in {0/Separability/{class distances,\\dimension control,\\LORO accuracy}, 1/Rotational dynamics/{rotation separability,\\stability of crops}, 2/Cross-continental transferability/{US$\to$India, India$\to$US,\\few target labels}} {
  \node[src, minimum height=1.35cm] (e\k) at (10.1,{-1.9-2.0*\k}) {{\bfseries \t}\\[1pt]\l};
}
\node[sec] at ($(p3.south east)+(-0.12,0.08)$) {Sec.\ 2.3--2.6};
% ---- 4 Controls and 5 Outputs
\node[panel, minimum width=3.4cm, minimum height=4.0cm] (p4) at (12.6,0) {};
\node[tab] at (p4.north) {4\ Controls};
\node[ctrl, anchor=north west] at (12.75,-0.95) {$\bullet$ matched dimensions\\$\bullet$ label permutation\\$\bullet$ crop identity removed\\$\bullet$ LORO hold-out\\$\bullet$ blocked reference\\$\bullet$ prior, target labels};
\node[sec] at ($(p4.south east)+(-0.12,0.08)$) {Sec.\ 2.4--2.6};
\node[panel, minimum width=3.4cm, minimum height=2.4cm] (p5) at (12.6,-5.0) {};
\node[tab] at (p5.north) {5\ Findings};
\node[ctrl, anchor=north west] at (12.75,-5.5) {\scriptsize\textbf{C1}\ Separability\\\textbf{C2}\ Rotational dynamics\\\textbf{C3}\ Cross-continental\\\phantom{\textbf{C3}}\ transferability};
\draw[flow] (14.3,-4.04) -- (14.3,-4.62);
% ---- horizontal flow arrows (clearly visible, 0.8 cm gaps)
\draw[flow] (3.52,-3.7) -- (4.08,-3.7);
\draw[flow] (7.72,-3.7) -- (8.28,-3.7);
\draw[flow] (11.92,-2.25) -- (12.48,-2.25);
\end{tikzpicture}}}
\caption{\textbf{Evaluation framework.} Embeddings, three engineered feature sets and their combination are paired with the crop grown in each year and evaluated for (i)~separability with dimension control, (ii)~rotational dynamics and (iii)~cross-continental transferability, with controls for dimensionality, crop identity and spatial hold-out. All sets are scored on identical samples. The findings C1--C3 correspond to the three contributions listed in the Introduction.}
\label{fig:workflow}
\end{figure}

\FloatBarrier
\subsection{Study Areas and Data}

Four datasets are used. They are (i) AlphaEarth embeddings \cite{brown2025alphaearth}, (ii) Sentinel-2 Multispectral Instrument (MSI) Level-2A imagery (atmospherically corrected surface reflectance), (iii) USDA Cropland Data Layer (CDL) \cite{boryan2011monitoring}, and (iv) the AgriFieldNet dataset for India \cite{agrifieldnet2022}. Raster inputs were resampled to a common 10~m grid (the CDL is natively 30~m). The CDL is itself a classification product, so the US crop labels carry its mapping errors, which are larger for specialty crops such as the orchard and vegetable crops of Fresno. It is nevertheless the standard source of year-by-year US crop sequences \cite{hunt2024crop}. AlphaEarth's training sources include land-cover records, and whether they overlap with the CDL was not tested (Section~\ref{sec:interp}). The AgriFieldNet labels are field-surveyed \cite{agrifieldnet2022}. The Sentinel-2 inputs are cloud-masked and restricted to the crop growing season (day of year, DOY, 90--300 for CONUS; the rabi season, 1~November to 30~April, for India).

The AlphaEarth embeddings are annual layers and are not restricted to the growing season, so each year's embedding is the layer for that calendar year. For the Indian fields, whose rabi 2021 crop was sown in late 2020 and harvested by April 2021, this is the 2021 layer, which covers the second half of the crop season and the months after it. The engineered features are averaged over the field polygon, whereas each embedding vector is pooled over a square window centered on a representative point inside the field ($320\times320$~m for the US patches and $160\times160$~m, 2.56~ha, for the Indian fields).

\begin{figure}[!ht]
\centering
\includegraphics[width=0.85\textwidth,height=0.5\textheight,keepaspectratio]{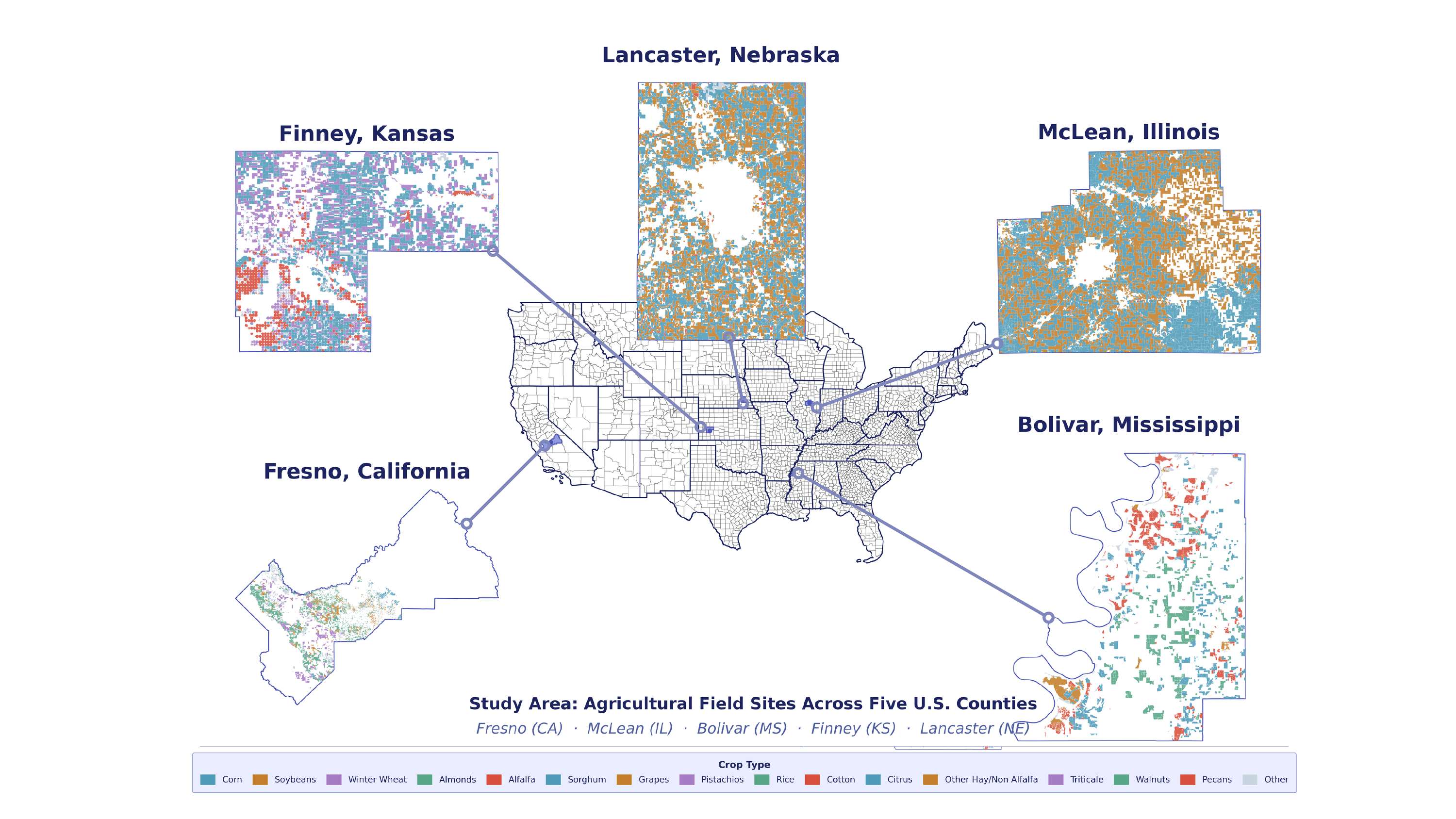}
\caption{\textbf{Study regions.} Five US counties: Fresno (CA), Finney (KS), Lancaster (NE), McLean (IL) and Bolivar (MS). Map lines are indicative and do not imply accepted national boundaries. The counties span contrasting cropping systems.}
\label{fig:conus_map}
\end{figure}

Five CONUS counties (Fig.~\ref{fig:conus_map}; Table~\ref{tab:conus_summary}) are selected to represent diverse agro-ecological conditions, crop management practices, and landscape structures. Fresno (CA) represents highly diverse, specialty-crop systems; McLean (IL) represents homogeneous, large-scale corn--soybean systems; Bolivar (MS) captures double-cropping and hydrologically dynamic conditions; Lancaster (NE) represents transitional agro-climatic conditions with mixed irrigation; and Finney (KS) reflects semi-arid systems with wheat--sorghum--fallow rotations. The counties differ in crop diversity (9--21 crop classes), field size (county mean 76--260~ha) and climate.

Homogeneous crop patches in the CONUS are delineated by applying 8-neighbor Connected-Component Analysis (CCA) \cite{rosenfeld1966sequential, he2017connected}
to annual CDL rasters, grouping contiguous pixels of identical crop-class labels
into candidate field objects. Unlike GEOBIA, which segments raw imagery to discover
field boundaries \cite{blaschke2010object, blaschke2014geographic}, CCA operates directly
on pre-classified label rasters, which yields spatially contiguous, label-homogeneous agricultural patches. CCA is used because the aim is to build temporally consistent sampling units for representation evaluation and rotation analysis, not to extract precise field boundaries \cite{yan2016conterminous, hunt2024crop, upcott2023new}. Because CCA joins adjacent pixels of the same crop, a patch can contain several neighboring farm fields. Patch areas (Table~\ref{tab:conus_summary}; county means of 76--260~ha, median 36~ha over all patches, maximum 19,450~ha) exceed typical field sizes, and these label-homogeneous patches are referred to as field units (``fields'' for brevity) when the US sampling unit is described. The Indian units, in contrast, are surveyed field polygons. ``Field'' thus means a label-homogeneous patch in the US data and a surveyed polygon in the Indian data. Because patch area varies widely, results are also reported by area class (Supplementary Section F).

Patches are retained using minimum area thresholds ($>$10~ha for Fresno;
$>$5~ha elsewhere) and a temporal consistency criterion on the patch geometry. A patch must be present as a delineated agricultural object in at least six of the seven years, so that retained patches are persistent agricultural units
and not one-year classification artifacts \cite{waldner2020deep, persello2019delineation}.
The criterion constrains the patch boundary and not the crop grown, so patches whose CDL class changes between years are kept, because rotations are the object of the multi-year analysis. Each patch carries a single crop label, which equals the crop recorded for 2019 (the third year of the 2017--2023 sequence) for 99.6\% of patches. Because of rotation, the crop in other years often differs from this label: on average 40\% of a patch's years differ from it (52\% for Corn, 45\% for Soybeans, 66\% for Winter Wheat and 24\% for Alfalfa). The analyses reported here pair each year's features with the crop of that year, taken from the crop sequence, and the stored 2019 label is used only for field-level summaries (the Structural Deviation Index, SDI, and rotation separability) and for a comparison with the stored-label design (Supplementary Section B).

Five Fresno fields with no crop label and no valid embedding in any year are excluded, leaving 6,315 fields (Table~\ref{tab:conus_summary}).
The full 2017--2023 CDL time series is used to derive crop sequences and rotation
patterns \cite{boryan2011monitoring, hunt2024crop}, yielding diverse rotation classes
(Fig.~S6), with alternating corn--soybean as the
dominant pattern, followed by three-crop and multi-crop rotations. Perennial crops (almonds, pistachios, walnuts, grapes) form single-crop classes and not rotations, so these classes are distinguished by crop identity.

\begin{table}[!ht]
\centering
\caption{Summary of the 2017--2023 CONUS dataset (GEOID: county FIPS code; Crops: crop classes in the CDL series). Counties differ widely in patch size and crop diversity.}
\label{tab:conus_summary}
\scriptsize
\begin{tabular}{lrrrrc}
\toprule
\textbf{County} & \textbf{GEOID} & \textbf{Fields} & \textbf{Crops} & \textbf{Mean Area (ha)} & \textbf{Area Range (ha)} \\
\midrule
Fresno    & 06019 & 1,748 & 21 & 128.2 & 10.1--11,252.0 \\
McLean    & 17113 &   929 & 10 & 259.7 &  5.0--19,449.9 \\
Finney    & 20055 & 1,759 & 16 & 103.2 &  5.0--8,130.3  \\
Bolivar   & 28011 &   580 &  9 &  76.4 &  5.0--2,090.4  \\
Lancaster & 31109 & 1,299 &  9 & 114.8 &  5.0--5,919.6  \\
\bottomrule
\end{tabular}
\end{table}

For India, the AgriFieldNet dataset \cite{agrifieldnet2022} is processed by applying a 5~m interior buffer to reduce mixed-pixel effects and boundary noise prevalent in smallholder systems \cite{persello2019delineation}, which leaves 3,890 fields after preprocessing. The dataset covers 12 crop classes: wheat (52.0\% of fields), mustard (25.3\%), maize (7.5\%), sugarcane (4.2\%), rice (3.3\%), lentil (2.6\%), gram (chickpea, 1.5\%), garlic (1.2\%), potato (1.1\%), green pea (0.6\%), berseem (Egyptian clover, 0.4\%) and coriander (0.4\%). Of the fields, 94.0\% lie in Uttar Pradesh, 3.1\% in Odisha, 2.9\% in Madhya Pradesh and one field in a fourth region. Field area has a mean of 0.255~ha and a median of 0.147~ha, and 88\% of fields are under 0.5~ha.

\FloatBarrier
\subsection{Feature Sets and Data Properties}

Four base feature sets are used, one vector per field (the embedding extraction is described above):
\begin{enumerate}
    \item \textbf{Latent Embeddings:} 64-dimensional vectors capturing multi-sensor spatio-temporal context \cite{brown2025alphaearth, ma2025harvesting}.
    \item \textbf{Spectral Reflectance:} Median growing-season reflectance for Sentinel-2 bands (B2, B3, B4, B8, B11, B12), from Level-2A surface reflectance (scaled to 0--1 in the US extraction; see the note on Indian units below). Cloud, shadow and snow pixels are removed with the scene classification layer, and the bands are composited by the median over day 90--300 and averaged over the field polygon at 10~m.
    \item \textbf{NDVI Phenology:} Maximum, minimum, amplitude, and peak timing of NDVI time series \cite{zeng2020review, li2021high}.
    \item \textbf{Harmonic Features:} Seasonal phase, amplitude, and trigonometric coefficients derived from a 1-cycle harmonic model \cite{roy2020robust}.
\end{enumerate}
A fifth set, Combined, concatenates the embeddings with the three engineered sets (79 dimensions for the US data; 80 for the Indian data, which include an additional calendar-corrected phase feature).
The supplied tables also contain climate (CHIRPS, SMAP, ERA5-Land \cite{funk2015chirps, entekhabi2010smap, munozsabater2021era5land}), soil (SoilGrids \cite{hengl2017soilgrids}), terrain (SRTM \cite{farr2007shuttle}) and Sentinel-1 radar variables \cite{mcnairn2004application}. These groups are not part of the four feature sets of the main comparison and are analyzed in the Supplementary Material (Section J).
Two properties of the feature tables affect the analysis. First, the US harmonic table has a year column for four of the five counties but not for Fresno (12,236 rows, 1,748 fields), so the same harmonic values are used for every year of a Fresno field. The harmonic and combined results for Fresno therefore do not vary over time. Second, the Sentinel-2 band values of the Indian table are stored in different units from the US table (median B2 of 40.5 against 0.064). Within-region analyses are unaffected after standardization, and the cross-continental experiments are run both with scaling fitted on the training region and with each region standardized on its own statistics (Section~\ref{sec:cross_methods}).

\FloatBarrier
\subsection{Evaluation Design}

Validation uses field-year rows labeled with the crop of that year, taken from the seven-year crop sequence in the rotation map. Each year's features are therefore paired with the crop actually grown (the stored label equals the crop of 2019 and agrees with the crop of a given year in 59.5\% of rows). Only rows with complete data for every feature set are kept, so all sets are scored on identical samples; double-crop, fallow and catch-all classes are excluded, and a class is included in a county only if it has at least 50 rows there (39,610 rows, 6,241 fields and 19 classes).

Spatial generalization is assessed with leave-one-region-out (LORO) validation, in which each region is one county (the transfer experiments below use continents as regions) \cite{roberts2017cross}. The model is trained on four counties and tested on the fifth, which mimics mapping a county without local labels. The analysis uses the crops present in at least four of the five counties (Alfalfa, Corn, Soybeans and Winter Wheat; 24,730 rows from 4,580 fields). Each county is held out in turn, and a per-year scaler is fitted on the training rows only. Intervals for the county mean come from resampling whole fields within each held-out county, with paired differences computed on the same resamples. Spatial block cross-validation inside each county and the spatial structure of the errors are described in Supplementary Section A and reported in Supplementary Section D.

Macro F1, the unweighted mean of the per-class F1 scores, is averaged over the classes present in the held-out set (the field-size analysis in Supplementary Section F uses Weighted F1, which weights classes by size). Fresno has only two of the four crops (Alfalfa and Winter Wheat) and Bolivar three, so per-county scores average over different numbers of classes and are not directly comparable.

The main-text scores come from a Random Forest \cite{breiman2001random} (200 trees, class-balanced weights), run with multiple random seeds for the held-out-county and transfer scores (Supplementary Section I gives a seed check). The labeled-field experiments average five random draws, each with its own model seed, and the zero-shot corrections, the stored-label check and the comparison with other classifiers use seed 42. To test whether the rankings depend on the classifier, XGBoost \cite{chen2016xgboost}, LightGBM \cite{ke2017lightgbm} and a support vector machine (SVM) with a radial basis function kernel \cite{cortes1995support} are run on the same samples. L2-regularized logistic regression and a one-hidden-layer multi-layer perceptron (MLP) serve as linear and nonlinear probes. L1 and elastic-net logistic regression give the sparsity counts. Linear discriminant analysis (LDA) gives the out-of-fold probabilities for the area under the ROC curve (AUC) and the log-loss of the crop-pair scores. The settings of all models are listed in Supplementary Table~S10. The probes and other classifiers are reported in Supplementary Sections C and E.

\FloatBarrier
\subsection{Separability and Dimension Control}

All features are standardized using z-score normalization; for the trajectory and rotation measures below, which depend on feature scale, standardization is applied within each year.

\paragraph{Effective Dimensionality.}
Because separation measures depend on the number of dimensions, the effective dimensionality of each feature set is quantified. Representational structure is analyzed using principal component analysis (PCA) \cite{jolliffe2016principal} of each feature set. Let $\{\gamma_i\}_{i=1}^{d}$ denote the eigenvalues of the covariance matrix sorted in descending order. The effective dimensionality is quantified using the participation ratio (PR, $D_{\text{PR}}$):
\begin{equation}
D_{\text{PR}} = \frac{\left(\sum_i \gamma_i\right)^2}{\sum_i \gamma_i^2}
\end{equation}
The number of principal components required to explain a fraction $\alpha \in (0,1)$ of variance is denoted $n_\alpha$ (so $n_{95}$ is the number of components retaining 95\% of variance).

\paragraph{Class Separability.}
Complementary distribution-aware separability metrics \cite{cha2007comprehensive} are computed, namely the Fisher ratio \cite{fisher1936use}, Bhattacharyya distance \cite{bhattacharyya1946measure}, squared Mahalanobis distance $D^2$ \cite{richards2006remote}, and Jeffries--Matusita distance \cite{matusita1967notion}. The Bhattacharyya distance has also been used to estimate transferability of learned representations \cite{pandy2022transferability}. The Fisher ratio is the multivariate form $\mathbf{d}^{\top}\mathbf{S}_w^{-1}\mathbf{d}/\operatorname{tr}(\mathbf{S}_w)$, where $\mathbf{d}$ is the difference of the class means and $\mathbf{S}_w$ the summed class covariance. Distance-based metrics are computed per crop pair with covariance regularization ($10^{-6}$) and 500 bootstrap resamples of whole fields (all rows of a field are kept together, because a field contributes up to seven correlated rows). Pairwise AUC and log-loss use out-of-fold LDA probabilities. The Bhattacharyya distance combines a mean term and a covariance term. It assumes Gaussian classes, which is only approximate for bounded or circular variables such as NDVI and harmonic phase, so it is read together with the permutation baseline and the classification scores.

\paragraph{Dimension Control.}
Bhattacharyya and squared Mahalanobis distances are recomputed in three settings. The embeddings are reduced to 4, 6 and 15 principal components or to a random projection of six dimensions, and the three engineered sets are pooled (15 features). The grid of 4 and 6 components matches the 4--6 features of the single engineered sets and 15 matches the pooled sets. Each is compared with a label-permutation baseline at the same dimension (whole fields permuted), so the excess over the baseline is the separation not explained by dimensionality. The LORO run is also repeated with embeddings reduced to 5, 6 and 15 dimensions, by principal components fitted on the training counties or by Gaussian random projection, and with the three engineered sets pooled (15 features). Both grids include six and 15 dimensions, which are the settings compared across the two analyses.

\FloatBarrier
\subsection{Rotational Dynamics}

\paragraph{Temporal Dynamics and Stability.}
Temporal dynamics are modeled as multi-year embedding trajectories \cite{zhou2022crop, upcott2023new}, where $e_t \in \mathbb{R}^d$ is the field-level mean feature vector in growing season $t$ and $k$ is the number of years in the observation window (here $k = 7$ for the 2017--2023 CONUS dataset). Rotation separability is evaluated on field-level feature vectors averaged over the observation window. It is the ratio of the mean pairwise distance between rotation-class centroids to the square root of the mean within-class variance (trace of the class covariance), assessed via permutation testing (1,000 iterations). Because the field-level features are averaged across years, this measure reflects differences in crop composition among rotation classes, and crop identity is removed in a second analysis. At the field level, temporal consistency is quantified using the SDI. Let $T_i = \{e_{t-k+1}, \dots, e_t\}$ be the $k$-year feature trajectory of field $i$ (a sequence of annual embeddings), and let $T_{\text{centroid}}$ be the mean trajectory of all fields assigned to the same crop class (i.e.\ the element-wise mean over all fields in that class). The SDI measures how much an individual field's trajectory deviates from its class mean:
\begin{equation}
\text{SDI}_i = 1 - \frac{T_i \cdot T_{\text{centroid}}}{\|T_i\|\,\|T_{\text{centroid}}\|}
\end{equation}
where lower values indicate higher consistency with the class mean. SDI is measured against class means defined by the stored 2019 label, so it reflects both rotation and year-to-year variation and is used to compare feature sets. At the class level, inter-annual stability is measured via the cosine similarity between class centroids across pairs of years. Let $c_t, c_{t'} \in \mathbb{R}^d$ denote the class centroid feature vectors in years $t$ and $t'$, respectively; then centroid similarity is:
\begin{equation}
\text{Sim}(c_t, c_{t'}) = \frac{c_t \cdot c_{t'}}{\|c_t\|\|c_{t'}\|}
\end{equation}
with values close to 1 indicating that the class representation is stable across years.

Crop identity is removed from the rotation measure in two further ways. Crop means are subtracted from the field features and rotation classes are permuted within crops, and rotation classes are compared inside single crops (classes with at least 30 fields). Stability over time is also tested directly. Models trained on 2017--2021 in all counties are tested on 2022--2023 in the same counties and fields, with year-matched labels. Each test year is standardized on its own unlabeled rows, and a scaler fitted on all training rows gives the same lead for the embedding and combined sets. Leave-one-year-out validation (training on all other years) and forward chaining (training on the years before the test year) serve as further checks. All five feature sets are scored. 

\FloatBarrier
\subsection{Cross-continental Transferability}
\label{sec:cross_methods}
To evaluate cross-continental transferability, models trained on CONUS data are tested on the Indian dataset and vice versa. The experiments use the three crop classes common to both regions (maize, rice and wheat), a Random Forest, the 2021 US data and the year-matched US label. Three variants are run. In training-region scaling, the scaler is fitted on the training region and applied unchanged to the target region. In per-region standardization, each region is z-scored on its own mean and standard deviation, which removes unit and level differences. The third variant repeats per-region standardization without the calendar-day features (peak timing, harmonic phase and its cosine and sine). Within-region scores for the same classes are obtained by stratified five-fold cross-validation. CDL ``Corn'' and ``Winter Wheat'' are matched to maize and wheat. US and Indian wheat, and US and Indian rice, are each treated as one class, although their calendars and management differ.

Cross-region consistency (between counties and between continents) is evaluated using Centered Kernel Alignment (CKA) \cite{kornblith2019similarity, cortes2012algorithms} and Earth Mover's Distance (EMD) \cite{rubner2000earth}. CKA captures structural alignment of feature spaces; EMD reflects distributional similarity. Fields in different counties are not paired, so for each crop the larger sample is randomly subsampled to the size of the smaller one and linear CKA is computed on the unpaired rows. Linear CKA on unpaired samples is expected to increase with feature dimensionality, so CKA values are compared qualitatively and should be read against a shuffled-row baseline.

Several diagnostics examine the transfer gap. The transfer is run with the stored label and with the year-matched label. The within-region reference is also scored with 10~km spatial blocks. An expectation-maximization (EM) estimate of the target class prior (which assumes calibrated probabilities) is applied after per-region standardization, once with the raw probabilities and once after recalibrating them with a logistic regression fitted on out-of-fold source predictions. Five, 20 and 50 labeled target fields per class (five random draws) are added to the source training set or used alone as the training set. The remaining target fields are scored, so that the contribution of the source region can be judged against local labels alone. Row-normalized confusion matrices of the transfer runs show which classes the predictions fall into (Supplementary Table~S7).

\FloatBarrier
\section{Results}
\label{sec:results}

This section follows the three research questions (RQ1, Section~\ref{sec:sep_results}; RQ2, Section~\ref{sec:temporal}; RQ3, Section~\ref{sec:transfer}) and ends with a ranking of the feature sets by task (Section~\ref{sec:summary_table}). The classification and centroid analyses pair each year's features with the crop of that year, whereas the SDI and the rotation measures use the stored 2019 label. Analyses that do not bear directly on the three themes are given in the Supplementary Material (Tables S1--S10 and Figures S1--S7).

\FloatBarrier
\subsection{Separability and Dimension-Controlled Separability}
\label{sec:sep_results}

Embeddings differ from the engineered sets in effective dimensionality. All engineered sets had low effective dimensionality ($n_{95} \leq 4$; $\text{PR} \approx 1.6$--$2.7$), while embeddings showed $n_{95} = 21$ (US) and $n_{95} = 26$ (India, all 3,890 fields) with participation ratios of 7.86 and 9.60, consistent with variance spread across many linear directions. Distances that grow with the number of dimensions therefore need a control, which is applied in Section~\ref{sec:matdim}.

\FloatBarrier
\subsubsection{Separating Look-Alike Crops}
\label{sec:dist_sep}

Embeddings separated look-alike crops by large margins, and the larger number of dimensions does not by itself explain the gap (Section~\ref{sec:matdim}).

\begin{figure}[!ht]
    \centering
    \includegraphics[width=\columnwidth]{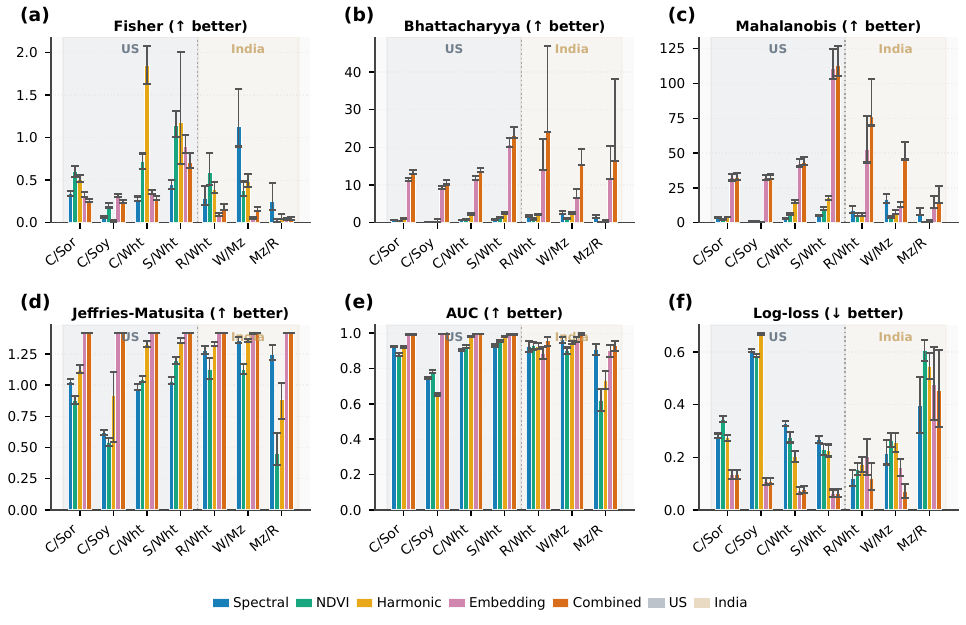}
    \caption{\textbf{Pairwise class separability} (year-matched labels), US (slate) and India (sand). Pairs: C/Sor corn--sorghum, C/Soy corn--soybeans, C/Wht corn--wheat, S/Wht soybeans--wheat, R/Wht rice--wheat, W/Mz wheat--maize, Mz/R maize--rice. (a)~Fisher ratio, (b)~Bhattacharyya, (c)~squared Mahalanobis, (d)~Jeffries--Matusita (maximum 1.41), (e)~AUC and (f)~log-loss from out-of-fold LDA; bars are point estimates, error bars 95\% field bootstrap (a--d). Embeddings and the combined set give the largest distribution-aware distances for every US pair.}
    \label{fig:separability}
\end{figure}

With each year's features paired to that year's crop, class separability depends strongly on the metric (Fig.~\ref{fig:separability}). For the look-alike pair Corn/Soybeans the multivariate Fisher ratio was highest for embeddings (0.32, against 0.02 for harmonic, 0.07 for spectral and 0.21 for NDVI features). For Corn/Wheat, however, harmonic features scored highest (1.85 against 0.35 for embeddings). The Fisher ratio, which is divided by the total within-class variance, therefore did not favor a single feature set across crop pairs. The distribution-aware distances were much larger for the embedding and combined sets in every US pair. The Bhattacharyya distance for Corn/Soybeans was 9.2 for embeddings (95\% interval 9.0--9.6) and 10.5 for the combined set, against 0.21 (0.20--0.23) for spectral, 0.16 (0.14--0.18) for NDVI and 0.55 (0.18--0.94) for harmonic features. AUC and log-loss from out-of-fold linear discriminant analysis show the same ordering for the US pairs (Corn/Soybeans AUC 0.997 for embeddings against 0.65--0.78 for the engineered sets). The ordering is less consistent for India, where spectral features match or exceed embeddings for rice/wheat and maize/rice (AUC 0.93 and 0.90 against 0.89 and 0.90). The Jeffries--Matusita distance is within 0.002 of its theoretical maximum of 1.414 for the embedding and combined sets in every pair, so at that ceiling it cannot rank them and does not indicate complete separation. Bhattacharyya and Mahalanobis distances increase with feature dimensionality, so the ordering Combined (79 dimensions) $>$ Embedding (64) $>$ engineered sets (4--6) partly reflects dimensionality, which Section~\ref{sec:matdim} controls for.

\FloatBarrier
\subsubsection{Accuracy on Counties Held Out of Training}
\label{sec:spatial}

Under LORO validation the Random Forest reached a mean Macro F1 of 0.83 with embeddings and 0.84 with the combined set, against 0.36--0.45 for the engineered sets (Fig.~S4a,b). The embedding advantage held in every held-out county (embeddings 0.73--0.92, the engineered sets 0.25--0.56), with the lowest embedding score in Fresno (0.73). It did not depend on Fresno, whose scores come from two classes only: without it the means were 0.84 for embeddings and 0.35--0.47 for the engineered sets. A bootstrap over fields within counties gave a 95\% interval of 0.81--0.85 for the embedding mean, and the leads over harmonic, spectral and NDVI features were 0.38, 0.47 and 0.47, each with an interval of about $\pm$0.02. Averaged over the four classifiers and paired by county, embeddings exceeded each engineered set in all five counties, by 0.37--0.45 on average (smallest county difference 0.17). The combined set was slightly ahead of the embeddings (mean difference 0.015 over four classifiers, with the embeddings ahead in two of five counties). With five counties the smallest attainable exact sign-test $p$-value is 0.06, so the counts are reported descriptively.

Pooled over the held-out counties, the errors were concentrated in specific crop pairs. Harmonic and NDVI features confused Corn and Soybeans (42\% of Corn predicted as Soybeans and 47\% of Soybeans as Corn for harmonic features, 37\% and 33\% for NDVI), against 3\% and 2\% for embeddings, so the Corn--Soybeans pair discriminated most strongly between feature sets, consistent with the distance results of Section~\ref{sec:dist_sep}. Further per-class errors are given in Supplementary Section D.

The ranking held with XGBoost, LightGBM and an SVM. The mean Macro F1 over the four classifiers was 0.82 for embeddings and 0.84 for the combined set, against 0.37 for spectral and NDVI and 0.45 for harmonic features. The combined set was level with the embeddings for the tree ensembles and ahead by 0.05 for the SVM (0.81 against 0.76; Table~S4). Linear and nonlinear probes and spatial block cross-validation inside each county gave the same ranking (Supplementary Sections C and D).

\FloatBarrier
\subsubsection{Dimension-Controlled Separability}
\label{sec:matdim}

\begin{table}[!ht]
\centering
\caption{Bhattacharyya distance at matched dimensionality (US pairs, year-matched labels): five feature sets, embeddings reduced to 4, 6 and 15 principal components (PCA on the two crops' rows), a random projection to six dimensions (mean of ten) and the three engineered sets pooled (15 features). The last row gives the field-permutation baseline for the 64-dimensional embeddings. Spec., spectral; Harm., harmonic; Emb., embedding; PCA-$k$, first $k$ principal components; Proj.-6, random projection to six dimensions; Eng. pooled, three engineered sets pooled. Embeddings stay ahead at six components, and the permutation baseline for the embeddings is small.}
\label{tab:matdim}
\scriptsize
\setlength{\tabcolsep}{3pt}
\begin{tabular}{lccccccccc}
\toprule
\textbf{Crop pair} & \textbf{Spec.} & \textbf{NDVI} & \textbf{Harm.} & \textbf{Emb.} & \textbf{PCA-4} & \textbf{PCA-6} & \textbf{PCA-15} & \textbf{Proj.-6} & \textbf{Eng. pooled} \\
\midrule
Corn/Sorghum & 0.75 & 0.49 & 0.99 & 11.12 & 3.73 & 3.92 & 4.90 & 1.90 & 2.12 \\
Corn/Soybeans & 0.21 & 0.16 & 0.55 & 9.16 & 3.58 & 3.83 & 4.60 & 1.94 & 1.01 \\
Corn/Winter Wheat & 0.65 & 0.79 & 2.13 & 11.47 & 4.11 & 4.25 & 5.42 & 2.04 & 3.05 \\
Soybeans/Winter Wheat & 0.76 & 1.25 & 2.47 & 20.74 & 8.19 & 9.60 & 11.20 & 3.48 & 3.84 \\
\midrule
Permutation baseline, Emb. & \multicolumn{9}{l}{0.31--0.57 for the 64-dimensional embeddings; at most 0.04 at 4--15 dimensions} \\
\bottomrule
\end{tabular}
\end{table}

Matching the dimensionality did not remove the embedding advantage (Table~\ref{tab:matdim}). With the embeddings reduced to six principal components, the Bhattacharyya distance for Corn/Soybeans was 3.8, against 0.2--0.6 for the single engineered sets and 1.0 for the three pooled (15 features), and the same held for the other three pairs. A random projection to six dimensions, which keeps no directions chosen for variance, gave 1.9 for Corn/Soybeans, still above the pooled engineered features. For the embeddings the label-permutation baseline was at most 0.04 at 4--15 dimensions and 0.3--0.6 at 64, so the excess over the baseline was almost the whole distance.

The same holds for accuracy on held-out counties (Table~\ref{tab:loro_dim}). With the embeddings reduced to their first six principal components, fitted on the training counties only, the LORO Macro F1 was 0.85, against 0.36--0.45 for the individual engineered sets and 0.49 for the three engineered sets pooled (15 features). A random projection to six dimensions performed worse (0.59) and one to 15 dimensions recovered most of the accuracy (0.74), so the useful information is spread across the embedding space, with most of it in the leading variance directions. Six components scored 0.85 against 0.83 for all 64, a difference smaller than the spread across counties (standard deviation 0.05--0.08), so six components kept the class information.

\begin{table}[!ht]
\centering
\caption{LORO Macro F1 at reduced dimension (Random Forest, year-matched labels, mean over five held-out counties; principal components fitted on the training counties only). Table~S2 gives the full grid. The embedding advantage persists at five to six principal components.}
\label{tab:loro_dim}
\scriptsize
\begin{tabular}{lccc}
\toprule
\textbf{Feature set} & \textbf{Dimensions} & \textbf{Macro F1} & \textbf{SD over counties} \\
\midrule
Embedding, all dimensions & 64 & 0.834 & 0.083 \\
Embedding, first 15 principal components & 15 & 0.832 & 0.079 \\
Embedding, first 6 principal components & 6 & 0.847 & 0.050 \\
Embedding, first 5 principal components & 5 & 0.775 & 0.143 \\
Embedding, random projection to 15 & 15 & 0.735 & 0.088 \\
Embedding, random projection to 6 & 6 & 0.589 & 0.179 \\
Engineered sets pooled & 15 & 0.493 & 0.118 \\
Harmonic (best single engineered set) & 5 & 0.453 & 0.114 \\
Spectral & 6 & 0.365 & 0.077 \\
NDVI & 4 & 0.361 & 0.098 \\
\bottomrule
\end{tabular}
\end{table}

\FloatBarrier
\subsection{Crop Rotations and Year-to-Year Stability}
\label{sec:temporal}

Embeddings gave the best separation of rotation classes overall, mostly because the classes differ in which crops they contain, and their lead shrinks once crop identity is removed.

\begin{figure}[!ht]
    \centering
    \includegraphics[width=\columnwidth]{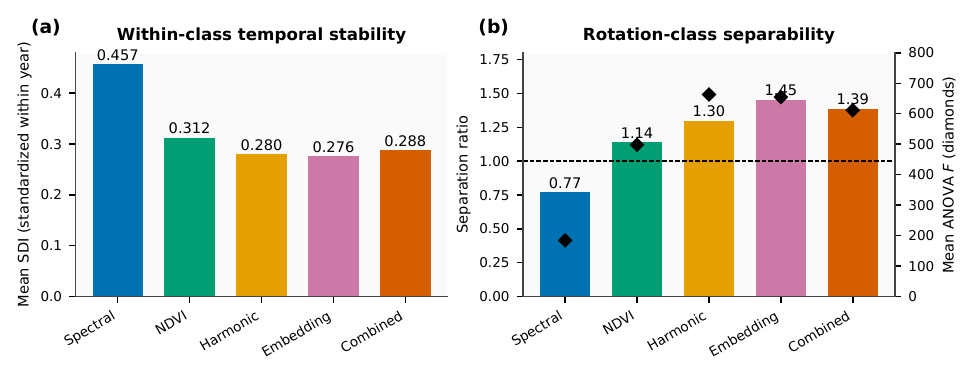}
    \caption{\textbf{Temporal stability and rotation separability} (US, standardized within year). (a)~Mean SDI (lower is more stable). (b)~Rotation separation ratio (bars) and mean analysis-of-variance (ANOVA) $F$ statistic (diamonds); permutation $p<0.001$ for all sets. SDI is similar for embedding, harmonic and combined features (0.276--0.288), and embeddings have the highest ratio (1.454).}
    \label{fig:rotation_dynamics}
\end{figure}

Embeddings had the highest rotation separation ratio (1.454; Fig.~\ref{fig:rotation_dynamics}b and Table~S5), followed by the combined (1.387), harmonic (1.298) and NDVI (1.139) sets, with spectral features lowest (0.770); all ratios differed from chance (permutation $p<0.001$). The embedding lead over the harmonic and combined sets was small (1.454 against 1.298 and 1.387). The ratio depends on feature scale, so only the standardized values are compared (unstandardized values in Supplementary Section G). Because the ratio is computed on year-averaged field features and rotation classes are defined by crop sequences, it is read together with the crop-residualized version below.

For temporal stability (Fig.~\ref{fig:rotation_dynamics}a), embedding, harmonic and combined features had nearly the same mean SDI (0.276, 0.280 and 0.288). NDVI was slightly higher (0.312), and spectral features were clearly higher (0.457), so spectral features deviated most from their class mean across years.

At the class level, with each year's rows labeled by that year's crop, the year-to-year cosine similarity of embedding centroids averaged 0.92 across the 22 crops with at least 100 fields (minimum 0.82), against 0.89 for harmonic, 0.83 for NDVI and 0.81 for spectral features. The main rotation crops were as stable as the rest (Corn 0.91, Soybeans 0.91, Winter Wheat 0.91, Sorghum 0.92). The stored-label analysis, which mixed several crops within each class, gave lower similarities (0.56--0.70 for the main crops) and therefore understated this stability.

Because the rotation ratio mostly reflects which crops a rotation class contains, two stricter versions were computed (Table~\ref{tab:rotation_within}). After removing each crop's mean from the field features and permuting rotation classes within crops, all five feature sets still separated the 15 rotation classes better than chance (permutation $p<0.001$). The ratios fell to 0.31--0.49, with embeddings highest (0.49), followed by the combined set (0.46) and harmonic features (0.40). For Corn (six rotation classes, 1,423 fields) embeddings led (1.32 against 0.93--1.12 for the engineered sets), and for Alfalfa they were also highest (0.83). For Soybeans, however, harmonic (0.99) and NDVI (0.90) features exceeded embeddings (0.76). For Winter Wheat, spectral and NDVI features were highest (0.48 and 0.46 against 0.40 for embeddings). The embedding lead in rotation separability survived the removal of crop identity overall (0.49 against 0.31--0.46) and was largest for Corn and Alfalfa. The differences between feature sets (0.07--0.16 for the standardized ratio, 0.03 after removing crop identity) are small, so they indicate a modest lead rather than a clear separation.

\begin{table}[!ht]
\centering
\caption{Rotation-class separation ratio beyond crop identity (US, standardized within year, field means). Crop-residualized: crop means removed, classes permuted within crops (15 classes, 6,202 fields). Within crop: classes with at least 30 fields. Best per row in bold; permutation $p<0.001$ except for spectral features in Soybeans (0.002) and Sorghum (0.018). Embeddings lead overall (0.49) but not in every crop.}
\label{tab:rotation_within}
\scriptsize
\begin{tabular}{lccccc}
\toprule
\textbf{Scope} & \textbf{Spectral} & \textbf{NDVI} & \textbf{Harmonic} & \textbf{Embedding} & \textbf{Combined} \\
\midrule
Crop-residualized (all) & 0.31 & 0.37 & 0.40 & \textbf{0.49} & 0.46 \\
Corn (6 classes) & 0.93 & 1.01 & 1.12 & \textbf{1.32} & 1.28 \\
Soybeans (4) & 0.29 & 0.90 & \textbf{0.99} & 0.76 & 0.76 \\
Alfalfa (4) & 0.61 & 0.67 & 0.43 & \textbf{0.83} & 0.80 \\
Winter Wheat (3) & \textbf{0.48} & 0.46 & 0.33 & 0.40 & 0.40 \\
Sorghum (2) & 0.29 & \textbf{0.48} & 0.43 & 0.38 & 0.38 \\
\bottomrule
\end{tabular}
\end{table}

Stability over time was also tested directly (Table~\ref{tab:stability}). Models trained on 2017--2021 and tested on the same fields in 2022--2023 reached a Macro F1 of 0.95 for embeddings and for the combined set, against 0.61--0.64 for the engineered sets. The lead of the embedding and combined sets did not depend on how the test years were standardized. It held in every county (embeddings 0.84--0.94,  the engineered sets 0.21--0.71). It also held in the year-by-year designs, where embeddings scored 0.89--0.97 in every test year and the engineered sets 0.17--0.79 (the low harmonic value is for 2017). The embedding lead over the best engineered set was about 0.3 in the main split and 0.17--0.33 in the single-year tests, so the class signatures of the embeddings were stable enough for a classifier to carry over to later years. The test keeps the same fields, so it measures stability over time and not generalization to new places. The scores of all sets are higher than under county hold-out (0.95 against 0.83 for embeddings, 0.64 against 0.45 for harmonic and 0.36 for NDVI features), which shows how much of the signal depends on place.

\begin{table}[!ht]
\centering
\caption{Stability over time (Random Forest, year-matched labels, same fields in training and test years; Macro F1). Ranges are over test years.  Table~S6 gives every test year.}
\label{tab:stability}
\scriptsize
\begin{tabular}{p{5.6cm}ccccc}
\toprule
\textbf{Design} & \textbf{Spectral} & \textbf{NDVI} & \textbf{Harmonic} & \textbf{Embedding} & \textbf{Combined} \\
\midrule
Train 2017--2021, test 2022--2023 (test years standardized on their own rows) & 0.61 & 0.64 & 0.64 & 0.95 & 0.95 \\
Same, scaler fitted on all training rows & 0.60 & 0.62 & 0.57 & 0.95 & 0.95 \\
Leave one year out (seven test years) & 0.58--0.72 & 0.60--0.70 & 0.17--0.79 & 0.89--0.97 & 0.89--0.97 \\
Train on earlier years only (test years 2020--2023) & 0.56--0.65 & 0.59--0.71 & 0.57--0.79 & 0.89--0.96 & 0.88--0.97 \\
\bottomrule
\end{tabular}
\end{table}

\FloatBarrier
\subsection{Cross-continental Transferability}
\label{sec:transfer}

Embedding-based models retained a smaller share of their accuracy on the other continent unless labeled target fields were added, while NDVI and harmonic models retained a larger share.

\FloatBarrier
\subsubsection{Distribution Match Between Continents}
\label{sec:crossreg}

Linear CKA did not distinguish the feature sets, because it was close to a shuffled-row baseline for every set, across continents (embeddings 0.014 against 0.017) and within the US (0.060 against 0.064). EMD gave a clearer ordering. Cross-continental EMD was lowest for NDVI (0.96) and harmonic features (1.09), higher for embeddings (1.46) and the combined set (1.48), and highest for spectral features (2.41).  All CKA and EMD values are given in Supplementary Section H (Fig.~S7).

\FloatBarrier
\subsubsection{Accuracy Within a Region and Across Continents}
\label{sec:classification}

\begin{figure}[!ht]
    \centering
    \includegraphics[width=\columnwidth]{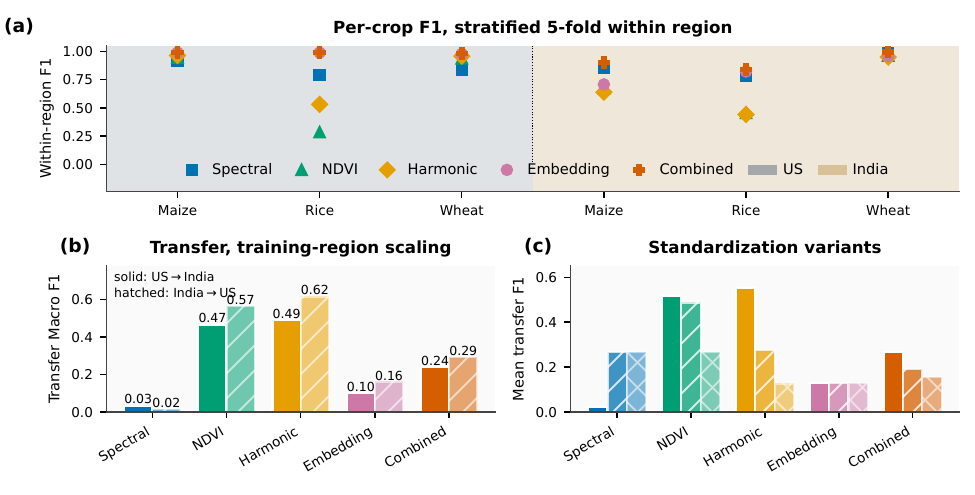}
    \caption{\textbf{Within-region and cross-continental classification} (Random Forest; maize, rice, wheat). (a)~Per-crop F1 under five-fold cross-validation inside each region (slate: US; sand: India). (b)~Transfer Macro F1 in both directions, scaling fitted on the training region. (c)~Mean transfer Macro F1 for training-region scaling, per-region standardization, and per-region standardization without calendar-day features. NDVI and harmonic features transfer best under training-region scaling, and embeddings are lowest or second lowest in every variant.}
    \label{fig:classification}
\end{figure}

Within a region, performance was high but varied with crop and region (Fig.~\ref{fig:classification}a). With the year-matched US label, embeddings reached a five-fold Macro F1 of 0.98 on the 2021 US fields (combined set 0.99, spectral 0.85, harmonic 0.82, NDVI 0.72). In India the combined set gave the highest Macro F1 (0.91), followed by spectral features (0.87), embeddings (0.83), NDVI (0.70) and harmonic features (0.68). Spectral features outperformed embeddings for several Indian crops (Supplementary Section H). Part of the US score reflects place. County identity alone, without any imagery, scored a Macro F1 of 0.71 in random folds (embeddings 0.99), because rice and wheat are concentrated in single counties (98\% of the US rice fields are in Bolivar and 96\% of the wheat fields in Finney).

Cross-continental transfer reversed the within-region ranking (Table~\ref{tab:transfer}; Fig.~\ref{fig:classification}b). Within regions, the combined and embedding sets scored highest in the US and the combined set in India. In transfer, harmonic and NDVI features scored highest. With scaling fitted on the training region, embeddings reached mean Macro F1 values of 0.16 (US$\to$India) and 0.22 (India$\to$US) over multiple seeds, which is 16\% and 28\% of their within-region scores, against 60\% and 93\% for harmonic and 68\% and 82\% for NDVI features. The embedding and combined scores varied between seeds (standard deviations 0.06--0.15), whereas those of the engineered sets were stable (0.00--0.02). A second multi-seed run gave lower embedding means (0.15 and 0.16) and a single run (seed 42) gave 0.10 and 0.16, so the transfer score of embeddings is itself variable, and the gap to the within-region score is large in every case. The combined set followed the embeddings rather than the engineered sets. Spectral features transferred lowest in both directions, consistent with the unit mismatch noted in Section~\ref{sec:methods}.

\begin{table}[!ht]
\centering
\caption{Within-region and cross-continental scores (Random Forest, maize, rice and wheat, year-matched US labels; Macro F1). Within-region scores are from five-fold cross-validation. Transfer uses scaling fitted on the training region and is given as the mean and standard deviation over multiple seeds. Retention is the mean transfer score divided by the within-region score of the source region scored with 10~km spatial blocks (Table~S8). Embeddings retain 16--28\% of their within-region accuracy.}
\label{tab:transfer}
\scriptsize
\setlength{\tabcolsep}{4pt}
\begin{tabular}{lcccccc}
\toprule
& \multicolumn{2}{c}{\textbf{Within region}} & \multicolumn{2}{c}{\textbf{Transfer, mean (SD)}} & \multicolumn{2}{c}{\textbf{Retention}} \\
\cmidrule(lr){2-3}\cmidrule(lr){4-5}\cmidrule(lr){6-7}
\textbf{Feature set} & US & India & US$\to$India & India$\to$US & US$\to$India & India$\to$US \\
\midrule
Spectral & 0.85 & 0.87 & 0.04 (0.02) & 0.02 (0.00) & 5\% & 2\% \\
NDVI & 0.72 & 0.70 & 0.47 (0.00) & 0.57 (0.02) & 68\% & 82\% \\
Harmonic & 0.82 & 0.68 & 0.48 (0.01) & 0.61 (0.01) & 60\% & 93\% \\
Embedding & 0.98 & 0.83 & 0.16 (0.10) & 0.22 (0.06) & 16\% & 28\% \\
Combined & 0.99 & 0.91 & 0.23 (0.11) & 0.36 (0.15) & 24\% & 41\% \\
\bottomrule
\end{tabular}
\end{table}

The confusion matrices of the transfer runs (Table~S7) clarify these scores. For embeddings the Random Forest assigned 96\% of the Indian fields to maize when trained on the US data (95\% of the maize fields, 96\% of the rice and 96\% of the wheat fields), and every US field to wheat when trained on the Indian data. The Macro F1 of 0.10 and 0.16 reflected an almost constant prediction in the target region rather than scattered errors. The spectral set behaved in the same way (all fields assigned to rice in both directions). NDVI and harmonic features kept some separation of maize and wheat (86\% and 93\% of Indian wheat fields were classified correctly by the US-trained NDVI and harmonic models, respectively), while rice, the smallest class, was almost never recovered by any set.

The ranking among the engineered sets depended on the experimental variant (Fig.~\ref{fig:classification}c). When each region was standardized on its own statistics, NDVI scored 0.45 and 0.53 (US$\to$India and India$\to$US), spectral features 0.20 and 0.34, and harmonic features 0.35 and 0.20. The harmonic advantage seen under training-region scaling therefore did not persist, and NDVI became the most transferable engineered set. When calendar-day features were also removed, NDVI fell to 0.26 and 0.27 and harmonic features to 0.13 and 0.12. Part of their performance therefore depended on region-specific timing. In every variant the embeddings remained among the lowest sets in both directions (0.10 and 0.16 with per-region standardization, against 0.20 and 0.34 for spectral features, which gain most from standardization).

Two further checks gave the same result (details in Supplementary Section H). With the stored 2019 label, harmonic features scored 0.41 and 0.42 and NDVI 0.35 and 0.41, both lower than before. Embeddings scored 0.09 and 0.15 against 0.10 and 0.16, so noise in the US label does not account for the embedding transfer gap. Spectral features were the exception for US$\to$India (0.30 against 0.03), where they remain below harmonic and NDVI features. With XGBoost, LightGBM and an SVM (Table~S4), embeddings were below NDVI in all eight classifier-direction cells and below harmonic features in seven of eight. Their mean over classifiers was 0.25 and 0.19, against 0.46 and 0.48 for NDVI and 0.42 and 0.61 for harmonic features. The embedding score depended on the classifier, from 0.10 (Random Forest) to 0.35 (LightGBM) for US$\to$India, so the size of the transfer gap was not a property of the embeddings alone.

\FloatBarrier
\subsubsection{Explaining the Transfer Gap and Restoring Accuracy}
\label{sec:diagnose}

The gap is not a simple shift in scale or class mix. Labeled fields from the new region raise the scores, but training on those fields alone does at least as well as adding them to the source data.

The retention figures divide the transfer score by a within-region score, which spatial autocorrelation can inflate. Scoring that reference with 10~km spatial blocks lowered it by at most 0.04 for every feature set (Table~S8; for example Indian embeddings from 0.83 to 0.79). The retention in Table~\ref{tab:transfer} uses the blocked reference, so it is not inflated by spatial autocorrelation, and the gap is not an artifact of the reference score. Retention also depends on the source score, so the absolute transfer scores in Table~\ref{tab:remedies} are the more direct comparison.

A simple offset or prior shift did not account for the gap either (Table~\ref{tab:remedies}). Per-region standardization left the embedding scores unchanged (0.10 and 0.16). A class-prior correction by expectation maximization did not help embeddings either. The estimated target prior concentrated on a single class (all maize for US$\to$India, all wheat for India$\to$US). The Random Forest assigned nearly all target fields to one class with high confidence, and the method requires calibrated probabilities. The correction left NDVI close to its earlier score for US$\to$India and lowered it for India$\to$US (0.46 and 0.41, from 0.45 and 0.53). It did not raise the embedding scores above those of NDVI. Recalibrating the probabilities with a logistic regression fitted on out-of-fold source predictions did not raise the embedding scores either (0.07 and 0.16). NDVI reached 0.46 and 0.58, so the gap originates in the embedding predictions and not in the prior estimate.

Adding labeled fields from the target region raised the embedding scores, but training on those fields alone did at least as well. With 50 labeled target fields per class added to the source data, embeddings reached 0.50 for US$\to$India and 0.92 for India$\to$US, from 0.10 and 0.16 without local labels. Trained on the same 50 target fields per class alone, embeddings reached 0.64 and 0.97 and the combined set 0.83 and 0.97. Target-only training was at least as good as adding the source data in 27 of the 30 combinations of feature set, direction and label count (5, 20 and 50 per class). There were three exceptions. NDVI for US$\to$India at 5 and 20 labeled target fields scored 0.51 and 0.57 against 0.54 and 0.58, and harmonic features for India$\to$US at 50 scored 0.70 against 0.72. For embeddings the gap was largest at 5 fields per class (0.50 and 0.96 with the target fields alone, against 0.36 and 0.58 with the source data added). The source-trained model added little once labeled target fields were available, and the gain comes mainly from the local labels and not from carrying the source model across. Where a class had fewer than 100 fields, at most half was labeled, so at most 23 of the 46 US rice fields were labeled.

\begin{table}[!ht]
\centering
\caption{Cross-continental transferability with adaptation steps (Macro F1, Random Forest, three common crops): EM estimate of the target class prior (raw and recalibrated), 5, 20 or 50 labeled target fields per class added to the source training set (``Source +'', mean of five draws), and the same labeled target fields used alone (``Target only''). Spec., spectral; Harm., harmonic; Emb., embedding; Comb., combined. Best per row and direction in bold. The labeled-field rows are means of five random draws, each with its own model seed, and the first four rows use seed 42; the multi-seed zero-shot scores are in Table~\ref{tab:transfer}. Standardization and prior correction do not help embeddings, and training on the target fields alone is at least as good as adding them to the source data in most cases.}
\label{tab:remedies}
\scriptsize
\setlength{\tabcolsep}{3pt}
\begin{tabular}{lccccccccccc}
\toprule
& \multicolumn{5}{c}{\textbf{US$\to$India}} & & \multicolumn{5}{c}{\textbf{India$\to$US}} \\
\cmidrule(lr){2-6}\cmidrule(lr){8-12}
\textbf{Condition} & \textbf{Spec.} & \textbf{NDVI} & \textbf{Harm.} & \textbf{Emb.} & \textbf{Comb.} & & \textbf{Spec.} & \textbf{NDVI} & \textbf{Harm.} & \textbf{Emb.} & \textbf{Comb.} \\
\midrule
Training-region scaling & 0.03 & 0.47 & \textbf{0.49} & 0.10 & 0.24 & & 0.02 & 0.57 & \textbf{0.62} & 0.16 & 0.29 \\
Per-region standardization & 0.20 & \textbf{0.45} & 0.35 & 0.10 & 0.11 & & 0.34 & \textbf{0.53} & 0.20 & 0.16 & 0.27 \\
Standardization + EM prior & 0.07 & \textbf{0.46} & 0.31 & 0.07 & 0.07 & & 0.16 & \textbf{0.41} & 0.16 & 0.16 & 0.16 \\
Standardization + EM prior, calibrated & 0.07 & \textbf{0.46} & 0.03 & 0.07 & 0.07 & & 0.29 & \textbf{0.58} & 0.16 & 0.16 & 0.16 \\
\midrule
Source + 5 fields per class & 0.36 & \textbf{0.54} & 0.45 & 0.36 & 0.49 & & 0.53 & 0.55 & 0.63 & 0.58 & \textbf{0.71} \\
Source + 20 fields per class & 0.42 & 0.58 & 0.50 & 0.42 & \textbf{0.62} & & 0.53 & 0.55 & 0.66 & 0.82 & \textbf{0.88} \\
Source + 50 fields per class & 0.47 & 0.57 & 0.51 & 0.50 & \textbf{0.68} & & 0.57 & 0.58 & 0.72 & 0.92 & \textbf{0.94} \\
\midrule
Target only, 5 fields per class & \textbf{0.70} & 0.51 & 0.55 & 0.50 & 0.63 & & 0.56 & 0.55 & 0.69 & \textbf{0.96} & 0.93 \\
Target only, 20 fields per class & 0.71 & 0.57 & 0.56 & 0.58 & \textbf{0.78} & & 0.64 & 0.61 & 0.68 & \textbf{0.97} & \textbf{0.97} \\
Target only, 50 fields per class & 0.72 & 0.61 & 0.57 & 0.64 & \textbf{0.83} & & 0.66 & 0.65 & 0.70 & \textbf{0.97} & \textbf{0.97} \\
\bottomrule
\end{tabular}
\end{table}

\subsection{Summary: Which Features for Which Task?}
\label{sec:summary_table}
\begin{table}[!ht]
\centering
\caption{Relative suitability of the feature sets by criterion. Ratings are relative within each row: \textbf{++} best; \textbf{+} above the median; \textbf{$\sim$} intermediate or no clear difference; $\mathbf{-}$ lowest or near-lowest. Embedding and Combined sets lead on separation, rotation and LORO accuracy, whereas NDVI and harmonic features lead on zero-shot cross-continental criteria and local labels reverse that ranking. Combined: embeddings plus the three engineered sets. n/a: not run.}
\label{tab:suitability}
\scriptsize
\setlength{\tabcolsep}{4pt}
\renewcommand{\arraystretch}{1.1}
\begin{threeparttable}
\begin{tabular}{p{7.4cm} >{\centering\arraybackslash}p{1.7cm} >{\centering\arraybackslash}p{1.4cm} >{\centering\arraybackslash}p{1.2cm} >{\centering\arraybackslash}p{1.5cm} >{\centering\arraybackslash}p{1.6cm}}
\toprule
\textbf{Evaluation Criterion} & \textbf{Embedding} & \textbf{Spectral} & \textbf{NDVI} & \textbf{Harmonic} & \textbf{Combined} \\
\midrule
\multicolumn{6}{l}{\textit{Geometric Separability}} \\
Mean-based separability, Fisher ratio (Corn/Soybeans) & ++ & $-$ & + & $-$ & + \\
Distribution-aware separation, Bhattacharyya$^{\dagger}$ & ++ & $-$ & $-$ & $-$ & ++ \\
Distribution-aware separation, squared Mahalanobis$^{\dagger}$ & ++ & $-$ & $-$ & $-$ & ++ \\
Accuracy on held-out counties (LORO, Random Forest Macro F1; ++ $\geq$0.8, $\sim$ 0.4--0.5, $-$ $<$0.4) & ++ & $-$ & $-$ & $\sim$ & ++ \\
Accuracy at reduced dimension (LORO; embeddings at 6 principal components 0.85, engineered sets pooled at 15 features 0.49; ++ $\geq$0.8, $-$ $<$0.5) & ++ & $-$ & $-$ & $-$ & n/a \\
\midrule
\multicolumn{6}{l}{\textit{Temporal Dynamics}} \\
Within-class temporal consistency (SDI, standardized; lower = more consistent; ++ $<$0.285, + 0.285--0.35, $-$ $>$0.35) & ++ & $-$ & + & ++ & + \\
Rotation-class separability (field-mean ratio, standardized; ++ $\geq$1.4, + 1.2--1.4, $\sim$ 1.0--1.2, $-$ $<$1.0) & ++ & $-$ & $\sim$ & + & + \\
Accuracy on later years (same fields, 2022--2023; ++ $\geq$0.8, + 0.5--0.8) & ++ & + & + & + & ++ \\
\midrule
\multicolumn{6}{l}{\textit{Cross-continental Transferability}} \\
Distributional alignment across regions (EMD, US$\to$India; lower = better) & $\sim$ & $-$ & ++ & + & $\sim$ \\
Cross-continental transferability (US$\to$India, training-region scaling) & $-$ & $-$ & + & ++ & $\sim$ \\
Cross-continental transferability (India$\to$US, training-region scaling) & $-$ & $-$ & + & ++ & $\sim$ \\
Training on 50 labeled target fields per class alone (mean of both directions; ++ $\geq$0.8, + 0.6--0.8) & ++ & + & + & + & ++ \\
\bottomrule
\end{tabular}
\begin{tablenotes}[flushleft]
\tiny
\item[$\dagger$] Depends on feature dimensionality (64 and 79 dimensions for Embedding and Combined against 4--6 for the engineered sets).

\item Cut-offs for the ++, + and $-$ ratings are given in the row labels where they apply. Cross-continental ratings use training-region scaling; under per-region standardization the engineered sets are nearly tied (Section~\ref{sec:classification}). 
\end{tablenotes}
\end{threeparttable}
\end{table}
Table~\ref{tab:suitability} summarizes the relative ranking of the feature sets across the evaluation criteria of the three contributions (C1--C3). Ratings are relative within each row and describe ranking rather than absolute adequacy.

\vspace{1.5\baselineskip}
\section{Discussion}
\label{sec:discussion}

\subsection{Contributions of the Study}

Beyond the accuracy comparison itself, the study offers three lessons for evaluating embeddings. The first concerns labels. Pairing each year's features with that year's crop raised the embedding LORO score from 0.51 to 0.83 (Supplementary Section B). An evaluation that stores one label per field would therefore have given embeddings a Macro F1 lower by 0.32 in a rotating system. The second concerns dimensionality. Reduced to six principal components, the embeddings still separated look-alike crops far better than the three engineered sets pooled (Corn/Soybeans Bhattacharyya distance 3.8 against 1.0). A six-dimensional random projection (1.9) also stayed ahead, so the gap is not due to the larger number of features. The third concerns what a source region contributes. Local labels alone (5, 20 or 50 per class) did at least as well as the same labels added to the source data in 27 of 30 comparisons, a result that only a target-only baseline can reveal.

\FloatBarrier
\subsection{Interpretation of the Findings}
\label{sec:interp}

We take the three research questions in turn.

RQ1: do embeddings separate crops better, and does the number of dimensions matter? Embeddings separated look-alike crops by much larger distances than the engineered features did (Corn/Soybeans Bhattacharyya distance 9.2 against 0.16--0.55). The advantage held at six principal components (3.8 against 1.0), under a six-dimensional random projection (1.9) and on counties held out of training, where the mean Macro F1 over four classifiers was 0.82 against 0.37--0.45. The combined set was slightly more accurate than the embeddings alone, and in India spectral features matched or beat embeddings for some crop pairs. The comparison is made against compact engineered summaries of one seasonal window, so richer baselines and fine-tuned embeddings are natural next steps. The US labels come from the Cropland Data Layer and the Indian labels from field surveys. The overlap between AlphaEarth's training sources and the Cropland Data Layer was not tested. Because class and place are entangled in the US data (county identity alone scored a Macro F1 of 0.71 in random folds), the held-out-county scores carry the main conclusions. Intervals are given for them and for the separability distances.

RQ2: do embeddings capture change across years and rotations? Embeddings kept stable class signatures and gave the highest rotation separability (ratio 1.454), with a lead that survived the removal of crop identity, mainly for Corn and Alfalfa. Within-class stability (SDI) was similar for embedding, harmonic and combined features. A classifier trained on 2017--2021 reached a Macro F1 of 0.95 on 2022--2023 against 0.61--0.64 for the engineered sets. Because the fields are shared, this shows stable signatures, and the lower held-out-county score of 0.83 indicates how much place adds. Rotation ratios and the SDI are deterministic computations and carry no intervals.

RQ3: do embedding-based models keep their accuracy in another region? The transfer experiments show where embeddings need local data. Embeddings retained 16--28\% of their within-region Macro F1, against 60--93\% for NDVI and harmonic features, and their predictions concentrated on a single class in the target region. A plausible explanation is that embeddings encode local landscape context that differs between the two regions. The gap was not explained by the reference score, the US label, per-region standardization or a simple prior shift, and its size varied with the classifier. Local labels recovered the accuracy quickly. Adding 50 labeled target fields per class raised the embedding scores from 0.10 and 0.16 to 0.50 and 0.92, and training on those fields alone reached 0.64 and 0.97, so a small local label set supplied most of the accuracy. Reduced accuracy of this kind is well documented when models move between regions \cite{tuia2016domain, hao2020transfer}. NDVI and harmonic features retained the most accuracy, although the harmonic advantage depends on how the regions are standardized and on the calendar-day features. For a new region, a few tens of labeled fields per class are an efficient route.

The transfer comparison uses the three crops the two regions share (maize, rice and wheat) and one rabi season in India, mainly in Uttar Pradesh (94\% of fields). Region is combined with window size (320~m in the US, 160~m in India), layer period, sampling unit and calendar, and these factors move together, so the transfer scores describe what happens when a model and its labels are reused across these two systems, which is the practical question.

\FloatBarrier
\subsection{Implications and Outlook}
\label{sec:implications}

For practitioners, the results point in three directions. Within a region where labels exist for several counties or years, annual embeddings work well. They gave the highest or second-highest accuracy on held-out counties with every classifier (level with the combined set for the tree ensembles and 0.05 below it for the SVM) and need no hand-engineered features. Each year's data should be paired with that year's crop, because the stored-label design gave the embeddings a Macro F1 lower by 0.32 (Supplementary Section B) and would change any comparison in rotating systems. And moving to a new region or cropping calendar calls for local labels. The source-trained model added little once labeled target fields were available, so the options are to label a few tens of fields per class and train locally, or to use NDVI-type features, which retained more accuracy when local labels are unavailable. Even the best zero-shot score (0.48 and 0.61 Macro F1) is far below what local training gave. Where fields are much smaller than the embedding window, as in Indian smallholder systems, spectral features matched or beat the embeddings within the Indian data.

The results show a consistent trade-off between local accuracy and transferability. Embeddings gain most from local training data: the LORO score rises with the number of training counties (0.49 with one county to 0.83 with four). In the transfer task, engineered features, NDVI in particular, transfer better than embeddings and are the safer choice when local labels are unavailable, although the harmonic advantage does not persist under per-region standardization. Table~\ref{tab:suitability} summarizes the trade-offs by task.

The main rankings were consistent across four classifiers and two validation designs.

\FloatBarrier
\section{Conclusions}
\label{sec:conclusion}

We asked how far ready-made AlphaEarth embeddings can replace engineered features for crop mapping, and how much the answer depends on how the test is run. The study contributes a year-matched, dimension-controlled evaluation design, a test of rotation signal beyond crop identity, and a target-only baseline for cross-continental transfer. It used 6,315 US fields over seven years and 3,890 Indian smallholder fields, with four classifiers and three further probes.

Within the US counties, embeddings, alone or combined with engineered features, were the most separable and the most accurate on counties held out of training (mean Macro F1 0.82 against 0.37--0.45 over four classifiers). The lead survived reduction to six principal components, so it is not an effect of dimensionality. The comparison is against compact engineered features, and the combined set was slightly more accurate than the embeddings alone.

Embeddings gave the highest rotation separability, with a lead that persisted once crop identity was removed. A model trained on 2017--2021 kept a clear lead on the same fields in 2022--2023 (Macro F1 0.95 against 0.61--0.64 for the engineered sets). The class signatures were therefore stable over time, a property the held-out-county scores then test for new places.

The cross-continental experiments show how much local data embeddings need. In our three-crop, two-region setting, embeddings retained 16--28\% of their within-region accuracy without local labels, against 60--93\% for harmonic and NDVI features, and 50 labeled target fields per class lifted them to 0.50 and 0.92. Training on those fields alone reached 0.64 and 0.97, so a small local label set supplies most of the accuracy.

Three design choices changed the results in rotating systems: pairing each year's features with that year's crop, reading class distances against a matched-dimensionality baseline, and removing crop identity before interpreting rotation measures. In practice, the results suggest using annual embeddings where labeled data from the same region exist, and relying on a few tens of local labels per class, or on simpler NDVI-type features, when mapping a new region. Next steps are fine-tuning on local labels, adding radar and climate inputs \cite{ienco2019combining}, comparing explicit domain adaptation \cite{tuia2016domain} with training on the few local labels alone, and repeating the comparison against richer engineered baselines.

\section*{Declaration of Competing Interest}
The authors declare that they have no known competing financial interests or personal relationships that could have appeared to influence the work reported in this paper.
\section*{CRediT Authorship Contribution Statement}
\textbf{Priyanka Goyal:} Conceptualization, Methodology, Software, Validation, Investigation, Visualization, Writing -- original draft.
\textbf{Maitreyee Sar:} Data curation, Formal analysis, Writing -- review and editing.
\textbf{Adway Mitra:} Conceptualization, Methodology, Supervision, Writing -- review and editing.
\textbf{Manjira Sinha:} Supervision, Validation, Writing -- review and editing.

\section*{Funding}
This research was partially supported by the Indo-German Science and Technology Centre (IGSTC) under the project IGSTC-03458:PARAM-VC, and the ``Niche Academic \& Research Programme Initiation'' under the Institute of Eminence (IoE) Grant, IIT Kharagpur.

\section*{Declaration of generative AI}
During the preparation of this work the authors used a large language model in order to improve the language and readability of the text. After using this tool, the authors reviewed and edited the content as needed and take full responsibility for the content of the publication.

\section*{Data Availability}
Data used in this study are described in Section~\ref{sec:methods}. The AlphaEarth embeddings (annual Satellite Embedding collection) are publicly available through Google Earth Engine and Google Cloud Storage. The AgriFieldNet dataset (CC BY 4.0) is available from Radiant Earth (doi:10.34911/rdnt.wu92p1). The USDA Cropland Data Layer (CDL) can be accessed through the USDA NASS CropScape portal, and Sentinel-2 Level-2A imagery is available via Google Earth Engine. 

\section*{Acknowledgements}
The authors thank the providers of the AlphaEarth embeddings, the Radiant Earth Foundation for the AgriFieldNet dataset, and the USDA National Agricultural Statistics Service for the Cropland Data Layer. Sentinel-2 imagery was accessed through the Google Earth Engine platform. 
\bibliographystyle{elsarticle-num}
\bibliography{references}

\begin{thebibliography}{10}
\expandafter\ifx\csname url\endcsname\relax
  \def\url#1{\texttt{#1}}\fi
\expandafter\ifx\csname urlprefix\endcsname\relax\def\urlprefix{URL }\fi
\expandafter\ifx\csname href\endcsname\relax
  \def\href#1#2{#2} \def\path#1{#1}\fi

\bibitem{un2015agenda}
{United Nations}, \href{https://sdgs.un.org/2030agenda}{Transforming our world:
  the 2030 agenda for sustainable development}, General Assembly resolution
  A/RES/70/1 (2015).
\newline\urlprefix\url{https://sdgs.un.org/2030agenda}

\bibitem{omia2023remote}
E.~Omia, H.~Bae, E.~Park, M.~S. Kim, I.~Baek, I.~Kabenge, B.-K. Cho, Remote
  sensing in field crop monitoring: A comprehensive review of sensor systems,
  data analyses and recent advances, Remote Sensing 15~(2) (2023) 354.
\newblock \href {https://doi.org/10.3390/rs15020354}
  {\path{doi:10.3390/rs15020354}}.

\bibitem{boryan2011monitoring}
C.~Boryan, Z.~Yang, R.~Mueller, M.~Craig, Monitoring {US} agriculture: the {US}
  {D}epartment of {A}griculture, {N}ational {A}gricultural {S}tatistics
  {S}ervice, {C}ropland {D}ata {L}ayer program, Geocarto International 26~(5)
  (2011) 341--358.
\newblock \href {https://doi.org/10.1080/10106049.2011.562309}
  {\path{doi:10.1080/10106049.2011.562309}}.

\bibitem{upcott2023new}
E.~V. Upcott, P.~A. Henrys, J.~W. Redhead, S.~G. Jarvis, R.~F. Pywell, A new
  approach to characterising and predicting crop rotations using national-scale
  annual crop maps, Science of the Total Environment 860 (2023) 160471.
\newblock \href {https://doi.org/10.1016/j.scitotenv.2022.160471}
  {\path{doi:10.1016/j.scitotenv.2022.160471}}.

\bibitem{zeng2020review}
L.~Zeng, B.~D. Wardlow, D.~Xiang, S.~Hu, D.~Li, A review of vegetation
  phenological metrics extraction using time-series, multispectral satellite
  data, Remote Sensing of Environment 237 (2020) 111511.
\newblock \href {https://doi.org/10.1016/j.rse.2019.111511}
  {\path{doi:10.1016/j.rse.2019.111511}}.

\bibitem{li2021high}
S.~Li, L.~Xu, Y.~Jing, H.~Yin, X.~Li, X.~Guan, High-quality vegetation index
  product generation: A review of ndvi time series reconstruction techniques,
  International Journal of Applied Earth Observation and Geoinformation 105
  (2021) 102640.
\newblock \href {https://doi.org/10.1016/j.jag.2021.102640}
  {\path{doi:10.1016/j.jag.2021.102640}}.

\bibitem{matsushita2007sensitivity}
B.~Matsushita, W.~Yang, J.~Chen, Y.~Onda, G.~Qiu, Sensitivity of the enhanced
  vegetation index (evi) and normalized difference vegetation index (ndvi) to
  topographic effects: a case study in high-density cypress forest, Sensors
  7~(11) (2007) 2636--2651.
\newblock \href {https://doi.org/10.3390/s7112636}
  {\path{doi:10.3390/s7112636}}.

\bibitem{huete2002overview}
A.~Huete, K.~Didan, T.~Miura, E.~P. Rodriguez, X.~Gao, L.~G. Ferreira, Overview
  of the radiometric and biophysical performance of the modis vegetation
  indices, Remote sensing of environment 83~(1-2) (2002) 195--213.
\newblock \href {https://doi.org/10.1016/s0034-4257(02)00096-2}
  {\path{doi:10.1016/s0034-4257(02)00096-2}}.

\bibitem{blaschke2010object}
T.~Blaschke, Object based image analysis for remote sensing, ISPRS journal of
  photogrammetry and remote sensing 65~(1) (2010) 2--16.
\newblock \href {https://doi.org/10.1016/j.isprsjprs.2009.06.004}
  {\path{doi:10.1016/j.isprsjprs.2009.06.004}}.

\bibitem{blaschke2014geographic}
T.~Blaschke, G.~J. Hay, M.~Kelly, S.~Lang, P.~Hofmann, E.~Addink, R.~Q.
  Feitosa, F.~Van~der Meer, H.~Van~der Werff, F.~Van~Coillie, et~al.,
  Geographic object-based image analysis--towards a new paradigm, ISPRS journal
  of photogrammetry and remote sensing 87 (2014) 180--191.
\newblock \href {https://doi.org/10.1016/j.isprsjprs.2013.09.014}
  {\path{doi:10.1016/j.isprsjprs.2013.09.014}}.

\bibitem{do2019detailed}
H.~do~Nascimento~Bendini, L.~M.~G. Fonseca, M.~Schwieder, T.~S. K{\o}rting,
  P.~Rufin, I.~D.~A. Sanches, P.~J. Leitao, P.~Hostert, Detailed agricultural
  land classification in the brazilian cerrado based on phenological
  information from dense satellite image time series, International Journal of
  Applied Earth Observation and Geoinformation 82 (2019) 101872.
\newblock \href {https://doi.org/10.1016/j.jag.2019.05.005}
  {\path{doi:10.1016/j.jag.2019.05.005}}.

\bibitem{roy2020robust}
D.~Roy, L.~Yan, Robust landsat-based crop time series modelling, Remote Sensing
  of Environment 238 (2020) 110810.
\newblock \href {https://doi.org/10.1016/j.rse.2018.06.038}
  {\path{doi:10.1016/j.rse.2018.06.038}}.

\bibitem{zhou2022crop}
X.~Zhou, J.~Wang, Y.~He, B.~Shan, Crop classification and representative crop
  rotation identifying using statistical features of time-series sentinel-1 grd
  data, Remote Sensing 14~(20) (2022) 5116.
\newblock \href {https://doi.org/10.3390/rs14205116}
  {\path{doi:10.3390/rs14205116}}.

\bibitem{kerner2020rapid}
H.~Kerner, G.~Tseng, I.~Becker-Reshef, B.~Barker, B.~Munshell, M.~Paliyam,
  M.~Hosseini, Rapid response crop maps in data sparse regions, arXiv preprint
  arXiv:2006.16866 (2020).
\newblock \href {https://doi.org/10.48550/arXiv.2006.16866}
  {\path{doi:10.48550/arXiv.2006.16866}}.

\bibitem{bommasani2021opportunities}
R.~Bommasani, D.~A. Hudson, E.~Adeli, R.~Altman, S.~Arora, S.~von Arx, M.~S.
  Bernstein, J.~Bohg, A.~Bosselut, E.~Brunskill, et~al., On the opportunities
  and risks of foundation models, arXiv preprint arXiv:2108.07258 (2021).
\newblock \href {https://doi.org/10.48550/arXiv.2108.07258}
  {\path{doi:10.48550/arXiv.2108.07258}}.

\bibitem{xiao2025foundation}
A.~Xiao, W.~Xuan, J.~Wang, J.~Huang, D.~Tao, S.~Lu, N.~Yokoya, Foundation
  models for remote sensing and earth observation: A survey, IEEE Geoscience
  and Remote Sensing Magazine (2025).
\newblock \href {https://doi.org/10.1109/mgrs.2025.3576766}
  {\path{doi:10.1109/mgrs.2025.3576766}}.

\bibitem{cong2022satmae}
Y.~Cong, S.~Khanna, C.~Meng, P.~Liu, E.~Rozi, Y.~He, M.~Burke, D.~Lobell,
  S.~Ermon, Satmae: Pre-training transformers for temporal and multi-spectral
  satellite imagery, Advances in Neural Information Processing Systems 35
  (2022) 197--211.
\newblock \href {https://doi.org/10.52202/068431-0015}
  {\path{doi:10.52202/068431-0015}}.

\bibitem{szwarcman2025prithvi}
D.~Szwarcman, S.~Roy, P.~Fraccaro, O.~E. G{\'\i}slason, B.~Blumenstiel,
  R.~Ghosal, P.~H. De~Oliveira, J.~L. de~Sousa~Almeida, R.~Sedona, Y.~Kang,
  et~al., Prithvi-eo-2.0: A versatile multi-temporal foundation model for earth
  observation applications, IEEE Transactions on Geoscience and Remote Sensing
  64 (2026) 1--20.
\newblock \href {https://doi.org/10.1109/TGRS.2025.3642610}
  {\path{doi:10.1109/TGRS.2025.3642610}}.

\bibitem{brown2025alphaearth}
C.~F. Brown, M.~R. Kazmierski, V.~J. Pasquarella, W.~J. Rucklidge,
  M.~Samsikova, C.~Zhang, E.~Shelhamer, E.~Lahera, O.~Wiles, S.~Ilyushchenko,
  et~al., {AlphaEarth} foundations: An embedding field model for accurate and
  efficient global mapping from sparse label data, arXiv preprint
  arXiv:2507.22291 (2025).
\newblock \href {https://doi.org/10.48550/arXiv.2507.22291}
  {\path{doi:10.48550/arXiv.2507.22291}}.

\bibitem{lisaius2026embedding}
M.~C. Lisaius, S.~Keshav, A.~Blake, C.~Atzberger, Towards accessible
  smallholder crop classification in the groundnut basin of senegal,
  Environmental Research: Food Systems 3 (2026) 045006.
\newblock \href {https://doi.org/10.1088/2976-601x/aea0b8}
  {\path{doi:10.1088/2976-601x/aea0b8}}.

\bibitem{cai2026annual}
Y.~Cai, B.~Li, X.~Liu, X.~Jiang, Y.~Zhu, S.~Luo, Y.~Qin, S.~Xie, J.~Ye,
  H.~Shen, et~al., Annual 10-m high-resolution cropland maps for southeast asia
  since 2019 using {AlphaEarth} embeddings, Earth System Science Data
  Discussions 2026 (2026) 1--30.
\newblock \href {https://doi.org/10.5194/essd-2026-78}
  {\path{doi:10.5194/essd-2026-78}}.

\bibitem{yang2026evaluating}
L.~Yang, Y.~Gao, X.~Zhao, N.~Liang, R.~Ma, S.~Xi, X.~Zhang, R.~Wang, Evaluating
  the performance of {AlphaEarth} foundation embeddings for irrigated cropland
  mapping across regions and years, Remote Sensing 18~(7) (2026) 1065.
\newblock \href {https://doi.org/10.3390/rs18071065}
  {\path{doi:10.3390/rs18071065}}.

\bibitem{murakami2025within}
K.~Murakami, Within- and cross-regional crop classification for cool climate
  upland agriculture using {AlphaEarth}, agriRxiv (2025).
\newblock \href {https://doi.org/10.31220/agrirxiv.2025.00354}
  {\path{doi:10.31220/agrirxiv.2025.00354}}.

\bibitem{wang2026ai}
W.~Wang, H.~Guo, S.~He, F.~Qi, A.~Samat, D.~Wang, J.~Li, Ai-driven precision
  mapping of tea plantations using {AlphaEarth} foundations: A scalable
  solution for smart agricultural monitoring, Agriculture 16~(4) (2026) 412.
\newblock \href {https://doi.org/10.3390/agriculture16040412}
  {\path{doi:10.3390/agriculture16040412}}.

\bibitem{fang2026application}
J.~Fang, M.~Wu, Z.~Zhang, W.~Luo, Application of {AlphaEarth} {GeoFoundation}
  model for high-accuracy crop yield estimation, in: 2026 IEEE/CVF Winter
  Conference on Applications of Computer Vision Workshops (WACVW), 2026, pp.
  1465--1474.
\newblock \href {https://doi.org/10.1109/wacvw68408.2026.00158}
  {\path{doi:10.1109/wacvw68408.2026.00158}}.

\bibitem{ma2025harvesting}
Y.~Ma, Y.~Shen, A.~Swatantran, D.~B. Lobell, Harvesting {AlphaEarth}:
  Benchmarking the geospatial foundation model for agricultural downstream
  tasks, International Journal of Applied Earth Observation and Geoinformation
  149 (2026) 105258.
\newblock \href {https://doi.org/10.1016/j.jag.2026.105258}
  {\path{doi:10.1016/j.jag.2026.105258}}.

\bibitem{neophytides2026benchmarking}
S.~P. Neophytides, M.~Mavrovouniotis, D.~G. Hadjimitsis, Benchmarking
  {AlphaEarth} embeddings for land cover mapping in agricultural landscapes,
  in: IGARSS 2026 - 2026 IEEE International Geoscience and Remote Sensing
  Symposium, 2026, pp. 2527--2531.
\newblock \href {https://doi.org/10.1109/igarss60577.2026.11730789}
  {\path{doi:10.1109/igarss60577.2026.11730789}}.

\bibitem{shao2026benchmarking}
B.~Shao, L.~Di, C.~Zhang, H.~Li, Z.~Liu, H.~Li, X.~Wang, Y.~Wei, Benchmarking
  geospatial foundation models for field-scale crop green-up detection:
  {AlphaEarth} vs. prithvi-eo-2.0 with sentinel-1 sar fusion, in: 2026 14th
  International Conference on Agro-Geoinformatics (Agro-Geoinformatics), 2026,
  pp. 1--6.
\newblock \href
  {https://doi.org/10.1109/agro-geoinformatics70621.2026.11667499}
  {\path{doi:10.1109/agro-geoinformatics70621.2026.11667499}}.

\bibitem{khan2025evaluating}
H.~Khan, A.~Ahmad, Evaluating {AlphaEarth} foundation embeddings for pixel- and
  object-based land cover classification in {Google Earth Engine},
  Preprints.org (2025).
\newblock \href {https://doi.org/10.20944/preprints202511.2172.v1}
  {\path{doi:10.20944/preprints202511.2172.v1}}.

\bibitem{zhou2026improving}
Y.~Zhou, X.~Zhang, J.~Liao, L.~Liu, Improving global tree cover estimation by
  integrating {AlphaEarth} foundations embeddings and biome adaptive modeling,
  IEEE Transactions on Geoscience and Remote Sensing 64 (2026) 4420613.
\newblock \href {https://doi.org/10.1109/tgrs.2026.3748245}
  {\path{doi:10.1109/tgrs.2026.3748245}}.

\bibitem{wang2026evaluating}
Y.~Wang, X.~Zhang, G.~Gao, L.~Liu, Evaluating the efficacy of {AlphaEarth}
  embeddings for tree species mapping in complex terrain, IEEE Journal of
  Selected Topics in Applied Earth Observations and Remote Sensing 19 (2026)
  19719--19733.
\newblock \href {https://doi.org/10.1109/jstars.2026.3699640}
  {\path{doi:10.1109/jstars.2026.3699640}}.

\bibitem{karim2026alphaearth}
M.~F. Karim, P.~Washaya, M.~Potterf, R.~Modlinger, K.~Kuzelka, T.~Hlasny,
  {AlphaEarth} foundations provide superior accuracy in vegetation height
  mapping across severely disturbed forest areas, IEEE Journal of Selected
  Topics in Applied Earth Observations and Remote Sensing 19 (2026)
  29343--29354.
\newblock \href {https://doi.org/10.1109/jstars.2026.3727370}
  {\path{doi:10.1109/jstars.2026.3727370}}.

\bibitem{liu2026endtoend}
S.~Liu, End-to-end landsat land surface temperature reconstruction with
  {AlphaEarth} embeddings, IEEE Geoscience and Remote Sensing Letters (2026).
\newblock \href {https://doi.org/10.1109/lgrs.2026.3754195}
  {\path{doi:10.1109/lgrs.2026.3754195}}.

\bibitem{benavides2026what}
I.~F. Benavides-Mart{\'\i}nez, J.~Guthrie, J.~E. Arias, Y.~A.
  Garc{\'e}s-G{\'o}mez, {\'A}.~In{\'e}s Guzm{\'a}n-Alvis, C.~V.
  Portilla-Cabrera, S.~Mondal, A.~J. Allyn, A.~R. Ganguly, What on earth is
  {AlphaEarth}? exploring functional specialization and shared associations in
  geospatial embeddings for land-cover analysis, IEEE Transactions on
  Geoscience and Remote Sensing 64 (2026) 3003517, preprint: arXiv:2603.16911.
\newblock \href {https://doi.org/10.1109/tgrs.2026.3750104}
  {\path{doi:10.1109/tgrs.2026.3750104}}.

\bibitem{borocz2025first}
B.~B{\"o}r{\"o}cz, G.~Moln{\'a}r, First experiences with the {AlphaEarth}
  foundations model: A cosine distance-based evaluation, in: Proceedings of the
  20th International Symposium on Applied Informatics and Related Areas, Obuda
  University, 2025, pp. 41--46.
\newblock \href {https://doi.org/10.12700/ais.2025.008}
  {\path{doi:10.12700/ais.2025.008}}.

\bibitem{han2024bridging}
B.~Han, S.~Zhang, X.~Shi, M.~Reichstein, Bridging remote sensors with
  multisensor geospatial foundation models, in: Proceedings of the IEEE/CVF
  conference on computer vision and pattern recognition, 2024, pp.
  27852--27862.
\newblock \href {https://doi.org/10.1109/cvpr52733.2024.02631}
  {\path{doi:10.1109/cvpr52733.2024.02631}}.

\bibitem{nozawa2022evaluation}
K.~Nozawa, I.~Sato, Evaluation methods for representation learning: A survey.,
  in: IJCAI, 2022, pp. 5556--5563.
\newblock \href {https://doi.org/10.24963/ijcai.2022/776}
  {\path{doi:10.24963/ijcai.2022/776}}.

\bibitem{tuia2016domain}
D.~Tuia, C.~Persello, L.~Bruzzone, Domain adaptation for the classification of
  remote sensing data: An overview of recent advances, IEEE geoscience and
  remote sensing magazine 4~(2) (2016) 41--57.
\newblock \href {https://doi.org/10.1109/mgrs.2016.2548504}
  {\path{doi:10.1109/mgrs.2016.2548504}}.

\bibitem{tarasiou2023vits}
M.~Tarasiou, E.~Chavez, S.~Zafeiriou, Vits for sits: Vision transformers for
  satellite image time series, in: Proceedings of the IEEE/CVF Conference on
  Computer Vision and Pattern Recognition, 2023, pp. 10418--10428.
\newblock \href {https://doi.org/10.1109/cvpr52729.2023.01004}
  {\path{doi:10.1109/cvpr52729.2023.01004}}.

\bibitem{tseng2023lightweight}
G.~Tseng, R.~Cartuyvels, I.~Zvonkov, M.~Purohit, D.~Rolnick, H.~Kerner,
  Lightweight, pre-trained transformers for remote sensing timeseries, arXiv
  preprint arXiv:2304.14065 (2023).
\newblock \href {https://doi.org/10.48550/arXiv.2304.14065}
  {\path{doi:10.48550/arXiv.2304.14065}}.

\bibitem{hao2020transfer}
P.~Hao, L.~Di, C.~Zhang, L.~Guo, Transfer learning for crop classification with
  cropland data layer data (cdl) as training samples, Science of The Total
  Environment 733 (2020) 138869.
\newblock \href {https://doi.org/10.1016/j.scitotenv.2020.138869}
  {\path{doi:10.1016/j.scitotenv.2020.138869}}.

\bibitem{richards2006remote}
J.~A. Richards, X.~Jia, Remote sensing digital image analysis: an introduction,
  Springer, 2006.

\bibitem{benediktsson2015spectral}
J.~A. Benediktsson, P.~Ghamisi, Spectral-spatial classification of
  hyperspectral remote sensing images, Artech House, 2015.

\bibitem{foody2002status}
G.~M. Foody, Status of land cover classification accuracy assessment, Remote
  sensing of environment 80~(1) (2002) 185--201.
\newblock \href {https://doi.org/10.1016/s0034-4257(01)00295-4}
  {\path{doi:10.1016/s0034-4257(01)00295-4}}.

\bibitem{roberts2017cross}
D.~R. Roberts, V.~Bahn, S.~Ciuti, M.~S. Boyce, J.~Elith, G.~Guillera-Arroita,
  S.~Hauenstein, J.~J. Lahoz-Monfort, B.~Schr{\o}der, W.~Thuiller, et~al.,
  Cross-validation strategies for data with temporal, spatial, hierarchical, or
  phylogenetic structure, Ecography 40~(8) (2017) 913--929.
\newblock \href {https://doi.org/10.1111/ecog.02881}
  {\path{doi:10.1111/ecog.02881}}.

\bibitem{schratz2019hyperparameter}
P.~Schratz, J.~Muenchow, E.~Iturritxa, J.~Richter, A.~Brenning, Hyperparameter
  tuning and performance assessment of statistical and machine-learning
  algorithms using spatial data, Ecological Modelling 406 (2019) 109--120.
\newblock \href {https://doi.org/10.1016/j.ecolmodel.2019.06.002}
  {\path{doi:10.1016/j.ecolmodel.2019.06.002}}.

\bibitem{ferraciolli2019neglecting}
M.~A. Ferraciolli, F.~F. Bocca, L.~H.~A. Rodrigues, Neglecting spatial
  autocorrelation causes underestimation of the error of sugarcane yield
  models, Computers and Electronics in Agriculture 161 (2019) 233--240.
\newblock \href {https://doi.org/10.1016/j.compag.2018.09.003}
  {\path{doi:10.1016/j.compag.2018.09.003}}.

\bibitem{agrifieldnet2022}
{Radiant Earth Foundation}, {IDinsight}, Agrifieldnet competition dataset,
  version 1.0, Radiant MLHub dataset (2022).
\newblock \href {https://doi.org/10.34911/rdnt.wu92p1}
  {\path{doi:10.34911/rdnt.wu92p1}}.

\bibitem{hunt2024crop}
K.~A. Hunt, J.~Abernethy, P.~C. Beeson, M.~Bowman, S.~Wallander, R.~Williams,
  Crop sequence boundaries using usda national agricultural statistics service
  historic cropland data layers, Statistical Journal of the IAOS 40~(2) (2024)
  237--246.
\newblock \href {https://doi.org/10.3233/sji-230078}
  {\path{doi:10.3233/sji-230078}}.

\bibitem{rosenfeld1966sequential}
A.~Rosenfeld, J.~L. Pfaltz, Sequential operations in digital picture
  processing, Journal of the ACM (JACM) 13~(4) (1966) 471--494.
\newblock \href {https://doi.org/10.1145/321356.321357}
  {\path{doi:10.1145/321356.321357}}.

\bibitem{he2017connected}
L.~He, X.~Ren, Q.~Gao, X.~Zhao, B.~Yao, Y.~Chao, The connected-component
  labeling problem: A review of state-of-the-art algorithms, Pattern
  Recognition 70 (2017) 25--43.
\newblock \href {https://doi.org/10.1016/j.patcog.2017.04.018}
  {\path{doi:10.1016/j.patcog.2017.04.018}}.

\bibitem{yan2016conterminous}
L.~Yan, D.~P. Roy, Conterminous united states crop field size quantification
  from multi-temporal landsat data, Remote Sensing of Environment 172 (2016)
  67--86.
\newblock \href {https://doi.org/10.1016/j.rse.2015.10.034}
  {\path{doi:10.1016/j.rse.2015.10.034}}.

\bibitem{waldner2020deep}
F.~Waldner, F.~I. Diakogiannis, Deep learning on edge: Extracting field
  boundaries from satellite images with a convolutional neural network, Remote
  sensing of environment 245 (2020) 111741.
\newblock \href {https://doi.org/10.1016/j.rse.2020.111741}
  {\path{doi:10.1016/j.rse.2020.111741}}.

\bibitem{persello2019delineation}
C.~Persello, V.~A. Tolpekin, J.~R. Bergado, R.~A. De~By, Delineation of
  agricultural fields in smallholder farms from satellite images using fully
  convolutional networks and combinatorial grouping, Remote sensing of
  environment 231 (2019) 111253.
\newblock \href {https://doi.org/10.1016/j.rse.2019.111253}
  {\path{doi:10.1016/j.rse.2019.111253}}.

\bibitem{funk2015chirps}
C.~Funk, P.~Peterson, M.~Landsfeld, D.~Pedreros, J.~Verdin, S.~Shukla,
  G.~Husak, J.~Rowland, L.~Harrison, A.~Hoell, J.~Michaelsen, The climate
  hazards infrared precipitation with stations---a new environmental record for
  monitoring extremes, Scientific Data 2 (2015) 150066.
\newblock \href {https://doi.org/10.1038/sdata.2015.66}
  {\path{doi:10.1038/sdata.2015.66}}.

\bibitem{entekhabi2010smap}
D.~Entekhabi, E.~G. Njoku, P.~E. O'Neill, K.~H. Kellogg, W.~T. Crow, W.~N.
  Edelstein, J.~K. Entin, S.~D. Goodman, T.~J. Jackson, J.~Johnson, J.~Kimball,
  J.~R. Piepmeier, R.~D. Koster, N.~Martin, K.~C. McDonald, M.~Moghaddam,
  S.~Moran, R.~Reichle, J.~C. Shi, M.~W. Spencer, S.~W. Thurman, L.~Tsang,
  J.~Van~Zyl, The soil moisture active passive {(SMAP)} mission, Proceedings of
  the IEEE 98~(5) (2010) 704--716.
\newblock \href {https://doi.org/10.1109/JPROC.2010.2043918}
  {\path{doi:10.1109/JPROC.2010.2043918}}.

\bibitem{munozsabater2021era5land}
J.~Mu{\~n}oz-Sabater, E.~Dutra, A.~Agust{\'i}-Panareda, C.~Albergel,
  G.~Arduini, G.~Balsamo, S.~Boussetta, M.~Choulga, S.~Harrigan, H.~Hersbach,
  B.~Mart{\'e}ns, D.~G. Miralles, M.~Piles, N.~J. Rodr{\'i}guez-Fern{\'a}ndez,
  E.~Zsoter, C.~Buontempo, J.-N. Th{\'e}paut, {ERA5-Land}: A state-of-the-art
  global reanalysis dataset for land applications, Earth System Science Data
  13~(9) (2021) 4349--4383.
\newblock \href {https://doi.org/10.5194/essd-13-4349-2021}
  {\path{doi:10.5194/essd-13-4349-2021}}.

\bibitem{hengl2017soilgrids}
T.~Hengl, J.~Mendes~de Jesus, G.~B.~M. Heuvelink, M.~Ruiperez~Gonzalez,
  M.~Kilibarda, A.~Bla{\v{g}}oti{\'c}, W.~Shangguan, M.~N. Wright, X.~Geng,
  B.~Bauer-Marschallinger, M.~A. Guevara, R.~Vargas, R.~A. MacMillan, N.~H.
  Batjes, J.~G.~B. Leenaars, E.~Ribeiro, I.~Wheeler, S.~Mantel, B.~Kempen,
  {SoilGrids250m}: Global gridded soil information based on machine learning,
  PLOS ONE 12~(2) (2017) e0169748.
\newblock \href {https://doi.org/10.1371/journal.pone.0169748}
  {\path{doi:10.1371/journal.pone.0169748}}.

\bibitem{farr2007shuttle}
T.~G. Farr, P.~A. Rosen, E.~Caro, R.~Crippen, R.~Duren, S.~Hensley, M.~Kobrick,
  M.~Paller, E.~Rodriguez, L.~Roth, D.~Seal, S.~Shaffer, J.~Shimada, J.~Umland,
  M.~Werner, M.~Oskin, D.~Burbank, D.~Alsdorf, The shuttle radar topography
  mission, Reviews of Geophysics 45~(2) (2007) RG2004.
\newblock \href {https://doi.org/10.1029/2005RG000183}
  {\path{doi:10.1029/2005RG000183}}.

\bibitem{mcnairn2004application}
H.~McNairn, B.~Brisco, The application of {C}-band polarimetric {SAR} for
  agriculture: A review, Canadian Journal of Remote Sensing 30~(3) (2004)
  525--542.
\newblock \href {https://doi.org/10.5589/m03-069} {\path{doi:10.5589/m03-069}}.

\bibitem{breiman2001random}
L.~Breiman, Random forests, Machine learning 45~(1) (2001) 5--32.
\newblock \href {https://doi.org/10.1023/a:1010933404324}
  {\path{doi:10.1023/a:1010933404324}}.

\bibitem{chen2016xgboost}
T.~Chen, C.~Guestrin, Xgboost: A scalable tree boosting system, in: Proceedings
  of the 22nd acm sigkdd international conference on knowledge discovery and
  data mining, 2016, pp. 785--794.
\newblock \href {https://doi.org/10.1145/2939672.2939785}
  {\path{doi:10.1145/2939672.2939785}}.

\bibitem{ke2017lightgbm}
G.~Ke, Q.~Meng, T.~Finley, T.~Wang, W.~Chen, W.~Ma, Q.~Ye, T.-Y. Liu, Lightgbm:
  A highly efficient gradient boosting decision tree, Advances in neural
  information processing systems 30 (2017).

\bibitem{cortes1995support}
C.~Cortes, V.~Vapnik, Support-vector networks, Machine learning 20~(3) (1995)
  273--297.
\newblock \href {https://doi.org/10.1007/bf00994018}
  {\path{doi:10.1007/bf00994018}}.

\bibitem{jolliffe2016principal}
I.~T. Jolliffe, J.~Cadima, Principal component analysis: a review and recent
  developments, Philosophical transactions of the royal society A:
  Mathematical, Physical and Engineering Sciences 374~(2065) (2016) 20150202.
\newblock \href {https://doi.org/10.1098/rsta.2015.0202}
  {\path{doi:10.1098/rsta.2015.0202}}.

\bibitem{cha2007comprehensive}
S.-H. Cha, Comprehensive survey on distance/similarity measures between
  probability density functions, International Journal of Mathematical Models
  and Methods in Applied Sciences 1 (2007).

\bibitem{fisher1936use}
R.~A. Fisher, The use of multiple measurements in taxonomic problems, Annals of
  eugenics 7~(2) (1936) 179--188.
\newblock \href {https://doi.org/10.1111/j.1469-1809.1936.tb02137.x}
  {\path{doi:10.1111/j.1469-1809.1936.tb02137.x}}.

\bibitem{bhattacharyya1946measure}
A.~Bhattacharyya, On a measure of divergence between two multinomial
  populations, Sankhy{\=a}: the indian journal of statistics (1946) 401--406.

\bibitem{matusita1967notion}
K.~Matusita, On the notion of affinity of several distributions and some of its
  applications, Annals of the institute of Statistical Mathematics 19~(1)
  (1967) 181--192.
\newblock \href {https://doi.org/10.1007/bf02911675}
  {\path{doi:10.1007/bf02911675}}.

\bibitem{pandy2022transferability}
M.~P{\'a}ndy, A.~Agostinelli, J.~Uijlings, V.~Ferrari, T.~Mensink,
  Transferability estimation using bhattacharyya class separability, in:
  Proceedings of the IEEE/CVF Conference on Computer Vision and Pattern
  Recognition, 2022, pp. 9172--9182.
\newblock \href {https://doi.org/10.1109/cvpr52688.2022.00896}
  {\path{doi:10.1109/cvpr52688.2022.00896}}.

\bibitem{kornblith2019similarity}
S.~Kornblith, M.~Norouzi, H.~Lee, G.~Hinton, Similarity of neural network
  representations revisited, in: International conference on machine learning,
  PMlR, 2019, pp. 3519--3529.
\newblock \href {https://doi.org/10.48550/arXiv.1905.00414}
  {\path{doi:10.48550/arXiv.1905.00414}}.

\bibitem{cortes2012algorithms}
C.~Cortes, M.~Mohri, A.~Rostamizadeh, Algorithms for learning kernels based on
  centered alignment, The Journal of Machine Learning Research 13 (2012)
  795--828.

\bibitem{rubner2000earth}
Y.~Rubner, C.~Tomasi, L.~J. Guibas, The earth mover's distance as a metric for
  image retrieval, International journal of computer vision 40~(2) (2000)
  99--121.
\newblock \href {https://doi.org/10.1023/a:1026543900054}
  {\path{doi:10.1023/a:1026543900054}}.

\bibitem{ienco2019combining}
D.~Ienco, R.~Interdonato, R.~Gaetano, D.~H.~T. Minh, Combining sentinel-1 and
  sentinel-2 satellite image time series for land cover mapping via a
  multi-source deep learning architecture, ISPRS Journal of Photogrammetry and
  Remote Sensing 158 (2019) 11--22.
\newblock \href {https://doi.org/10.1016/j.isprsjprs.2019.09.016}
  {\path{doi:10.1016/j.isprsjprs.2019.09.016}}.

\end{thebibliography}

\clearpage
\includepdf[pages=-,pagecommand={}]{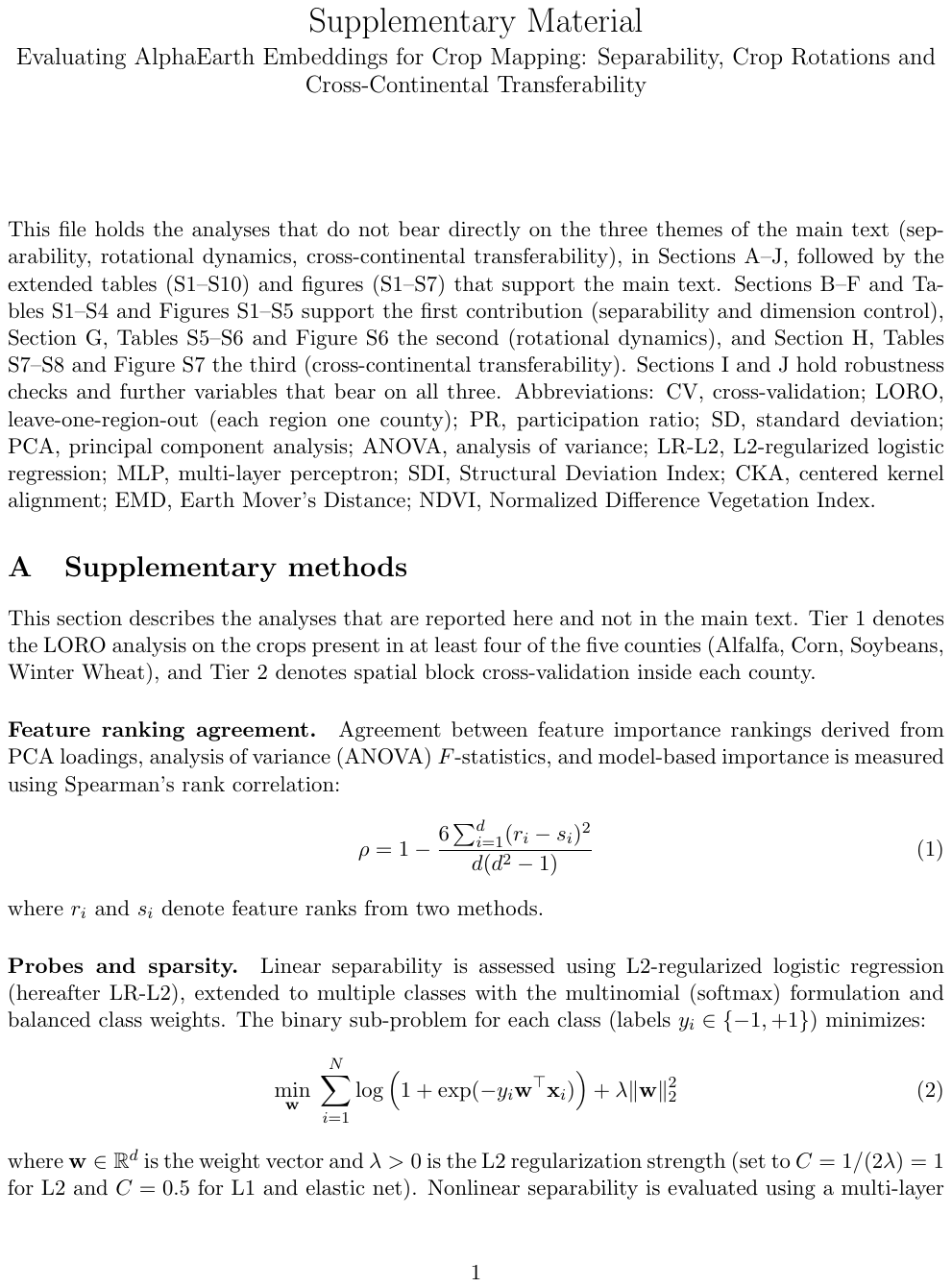}

\end{document}